\documentclass[12pt]{article}

\usepackage{amsmath,amsthm, amsfonts, amssymb, amsxtra, amsopn}
\usepackage{pgfplots}
\pgfplotsset{compat=1.3}
\usepackage{pgfplotstable}
\usepackage{graphicx,grffile}
\usepackage{multirow}
\usepackage{booktabs}
\usepackage{listings}
\usepackage{cmap}
\usepackage{colortbl}
\usepackage{adjustbox}
\usepackage{epsfig}

\usepackage[tableposition=top,font=small,skip=5pt]{caption}
\usepackage{subcaption}
\usepackage{makecell} 

\PassOptionsToPackage{hyphens}{url}

\usepackage{hyperref}
\hypersetup{colorlinks=true,linkcolor=black,citecolor=black,urlcolor=blue,filecolor=black}
\hypersetup{pdfpagemode=UseNone,pdfstartview=}

\usepackage{enumitem}
\setlist[itemize]{noitemsep, topsep=0pt}

\advance\oddsidemargin by -0.35in
\advance\textwidth by 0.7in

\advance\topmargin by -0.4in
\advance\textheight by 0.8in

\long\def\symbolfootnotetext[#1]#2{\begingroup%
\def\thefootnote{\fnsymbol{footnote}}\footnotetext[#1]{#2}\endgroup}

\title{On the Steganographic Capacity of Selected Learning Models}

\author{Rishit Agrawal\footnotemark[1]\ \ \ 
Kelvin Jou\footnotemark[1]\ \ \ 
Tanush Obili\footnotemark[1]\ \ \ 
Daksh Parikh\footnotemark[1]\ \ \ \\[0.75ex]
Samarth Prajapati\footnotemark[1]\ \ \ 
Yash Seth\footnotemark[1]\ \ \ 
Charan Sridhar\footnotemark[1]\ \ \ 
Nathan Zhang\footnotemark[1]\ \ \ \\[0.75ex]
Mark Stamp\footnotemark[1]\,\,\footnotemark[2]}

\begin{document}

\symbolfootnotetext[1]{Department of Computer Science, San Jose State University}
\symbolfootnotetext[2]{mark.stamp$@$sjsu.edu}

\maketitle

\abstract
Machine learning and deep learning models are potential vectors for various attack scenarios.
For example, previous research has shown that malware can be hidden in deep learning 
models. Hiding information in a learning model can be viewed as a form of steganography. 
In this research, we consider the general question of the steganographic capacity of 
learning models. Specifically, for a wide range of models,
we determine the number of low-order bits of the
trained parameters that can be overwritten, without
adversely affecting model performance. 
For each model considered, we graph the accuracy as 
a function of the number of low-order bits that have been overwritten, and for
selected models, we also analyze the steganographic capacity of individual layers.
The models that we test include the classic machine learning techniques
of Linear Regression (LR) and 
Support Vector Machine (SVM);
the popular general deep learning models of 
Multilayer Perceptron (MLP) and 
Convolutional Neural Network (CNN);
the highly-successful Recurrent Neural Network (RNN) architecture of
Long Short-Term Memory (LSTM);
the pre-trained transfer learning-based models
VGG16, DenseNet121, InceptionV3, and Xception; 
and, finally, an Auxiliary Classifier Generative Adversarial Network (ACGAN).
In all cases, we find that a majority of the bits of each trained parameter 
can be overwritten before the accuracy degrades. 
Of the models tested, the steganographic capacity
ranges from~7.04~KB for our LR experiments, to~44.74~MB
for InceptionV3. We discuss the implications of our results and consider 
possible avenues for further research.

\section{Introduction}\label{chap:Introduction}

The field of information hiding includes watermarking and steganography, which
use similar techniques, but for different purposes~\cite{Stamp_2021}. In digital watermarking, we want to 
hide information in a digital object, typically for the purpose of identifying the object. For example,
we might add a unique digital watermark to each copy of a confidential \texttt{pdf} files that we distribute.
Then, if a copy of the \texttt{pdf} is leaked to an unauthorized party, we could read the watermark to 
determine the source of the leak.

In contrast to watermarking, steganography consists of hiding information for
the purpose of communication. For example, if we want to communicate with
someone in a repressive country, we could hide information in a digital image of, say, a cat. 
If the recipient knows where and how to read the hidden information, we can
communicate on topics that would otherwise be censored.

Machine learning (ML), which can be considered as a subfield of Artificial Intelligence (AI), 
enables computers to learn important information from training data~\cite{Stamp_2022}. 
Today, ML models are widely used to deal with a vast array of problems, including speech recognition, 
image recognition, sentiment analysis, language translation, and malware detection, with new applications 
being constantly developed. Deep learning (DL) models are the subset of ML models that are based  
on neural networking techniques---they are ``deep'' in the sense of having multiple layers.

Machine learning models are of interest in the context of 
steganography for the following reasons.
\begin{enumerate} 
\item Machine learning models are rapidly becoming ubiquitous.
For example, learning-based voice-activated systems were used by more than~3.25 billion 
people in~2021~\cite{ML_Stat}.
\item The steganographic capacity of most
ML models is likely to be high. 
Models typically include a large number of weights or other trained parameters,
and learning models do not typically require high precision in their trained 
parameters. For example, the most popular algorithm used to train a Support Vector Machine (SVM) 
is Sequential Minimal Optimization (SMO), and the efficiency of this algorithm
relies on the fact that limited precision suffices~\cite{Stamp_2022}. 
As another example, in neural networking-based models,
many neurons tend to atrophy during training, with such weights contributing little to the 
trained model. By relying on such redundant neurons,
the authors of~\cite{EvilModel} show that they can hide~36.9~MB of malware
within a~178~MB AlexNet architecture, with only a~1\%\ degradation in performance. These 
changes do not affect the structure of the model and the embedded malware was
not detected by any of the anti-virus systems tested.
\item Machine learning models may be an ideal cover media for advanced malicious attacks. For example, 
in addition to simply embedding malware in a learning model,
it is conceivable that a specific predetermined input to the 
model could be used to trigger the embedded malware.
\end{enumerate}

As mentioned above, many learning models do not require
high precision in their trained parameters. Therefore, we propose to
measure the steganographic capacity of learning models by determining
the number of low-order bits of each weight that can be used for information hiding,
without adversely affecting the performance of a model. Specifically, 
we embed information in the~$n$ low-order bits of the weights of trained models, 
and graph the resulting model accuracy as a function of~$n$.
As our test case, we train models on a dataset that 
contains~10 different malware families, 
with a total of~15,356 samples. 

The remainder of the paper is organized as follows.
Section~\ref{chap:2} gives relevant background information. 
Section~\ref{chap:3} provides an overview the dataset used in our experiments,
and we outline our experimental design. Our results are
presented and discussed in Section~\ref{chap:4}.
Finally, Section~\ref{chap:5} gives our conclusions, and we discuss
potential topics for further research.

\section{Background}\label{chap:2}

In this section, we discuss relevant background topics. First, we discuss steganography,
then we briefly introduce the learning models that are used in this research. We 
conclude this section with a discussion of relevant related work.

\subsection{Steganography}

The word ``steganography'' is a combination of the Greek roots \textit{stegan\'{o}s}
and \textit{graphein}, which together translate as ``hidden writing''~\cite{Fiscutean_2021}. 
Thus, steganography consists of embedding 
information in a cover media~\cite{Stanger_2020}. 
In modern practice, digital steganography consists of concealing information 
within seemingly innocuous data, such as images, audio, video, or network communication, 
among other possibilities~\cite{Agarwal_2013}. 

We note in passing that cryptography protects 
a message by transforming it into an unintelligible format.
This is in contrast to steganography, where the goal is to hide the
fact that the communication represented by the hidden information
has even taken place. 
Steganography dates at least
to ancient Greece and it predates cryptography as a means 
of secret communication~\cite{Stamp_2021}. 

A simple example of a modern steganographic application consists of hiding 
information in the low order RGB bits of an uncompressed image file~\cite{Stamp_2021}. 
Since the RGB color scheme
uses a byte for each of the R (red), G (green), and B (blue) color components of each pixel, there 
are~$2^{24} > 16{,}000{,}000$ colors available. However, many of these color combination 
are indistinguishable to humans, and hence there are redundant bits in an uncompressed image
file that can be used for steganography. In particular,
the low-order RGB bits of each byte can be used to hide information, 
without perceptibly changing the image. Provided that the 
intended recipient knows which images are used for hiding information, and knows how to extract
the information, communication can take place between a sender and receiver, without
it being apparent that the hidden information has been communicated. The steganographic capacity of
an uncompressed image file is surprisingly large---in~\cite[Section~5.9.3]{Stamp_2021}
it is shown that the a \texttt{pdf} file containing the entire \textit{Alice's Adventures in Wonderland} book 
can be embedded in the low order RGB bits of a single image of Alice from the \textit{Alice} book itself.

The image-based steganographic technique in the previous paragraph is not robust, since
it is easy to disrupt the communication, without affecting the non-steganographic
use of such images. If a censor suspects that the low-order RGB bits of uncompressed image files are being
used for steganographic purposes, he can simply randomize the low-order bits of all such images.
The information would thus be lost from images that were being used for steganography, while
all other images would be unaffected in any perceptible way. 
Research in information hiding often focuses on creating 
more robust steganographic techniques.

The following three issues are relevant for a
steganographic technique.
 \begin{itemize}
\item Perceptual transparency --- 
A steganographic process should hide information
in a way that is imperceptible to human senses. This ensures that it is 
not obvious that the cover medium is being used for secret communication. 
\item Robustness --- 
As we noted in the case of image steganography discussed above, 
such a system may be more useful if it is robust.
\item Capacity --- 
The amount of information that can be hidden in the cover medium is the capacity.
The capacity of a steganographic technique depends on the redundancy in the cover media.
\end{itemize}

In this research, we are interested in the steganographic capacity of various learning  
models. Specifically, we hide information in the low-order bits of the trained parameters (i.e., weights) 
of selected learning models. While such a simple approach to information hiding 
is not robust, our work does provide a basis for designing more advanced techniques, 
with the analogy to image-based steganography being obvious.

\subsection{Learning Models}\label{sect:learningModels}

Machine learning (ML) and deep learning (DL) are tools used in the field of
artificial intelligence (AI). The general topic of ML deals with training ``machines'' 
to learn from data, and is often used for classification tasks. In our usage, DL is 
the subset of ML that uses Artificial Neural Networks (ANN), generally 
with multiple hidden layers, which is the source of the word ``deep'' in DL. 
Neural networking algorithms are designed to (loosely) mimic the structure 
of the human brain, and such models have proven to be very effective for solving 
complex problems such as image and speech recognition, 
natural language processing, and playing complex games at superhuman levels. 

ML enables computers to learn important information, and improve based on experience,
which saves humans from the inherently difficult task of extracting such information from
massive volumes of data~\cite{Stamp_2022}. A primary goal in the field of machine learning 
is to enable computers to learn, while requiring 
minimal human intervention or assistance~\cite{Selig_2022}.

As mentioned above, ML techniques are applied in a wide and growing range of fields.
ML techniques have become staples in the areas of data security, finance, healthcare, 
and so on. The subfield of DL has been particularly successful at dealing with such 
challenging problems as speech recognition, image classification, sentiment analysis, 
and language translation, among many others~\cite{Duggal_2022}. 

In recent years, DL models have achieved significant successes due to 
their ability to automatically extract complex patterns and representations from raw data 
without the requirement of extensive feature engineering. Through the process of training, DL models 
learn to recognize patterns and relationships in data, enabling them to often perform tasks 
at a higher level than had previously been achievable using ``classic'' ML models.

ML algorithms can be subdivided into 
supervised and unsupervised techniques. 
A supervised ML technique requires labeled data to train the model.
In contrast, unsupervised machine learning techniques can be applied to unlabeled data. 
In this paper, we only consider supervised learning techniques; specifically, we train
models to classify samples from a dataset containing~10 different malware families.

Next, we introduce each of the learning techniques that are employed in the experiments
in Section~\ref{chap:4}. Specifically, we discuss 
Logistic Regression (LR),
Support Vector Machine (SVM),
Multilayer Perceptron (MLP),
Convolutional Neural Network (CNN),
Long Short Term Memory (LSTM) models,
VGG16,
DenseNet121,
InceptionV3,
Xception,
and 
Auxiliary Classifier Generative Adversarial Network (ACGAN).


\subsubsection{Overview of Logistic Regression}

Logistic Regression (LR) models are a traditional machine learning algorithm, 
designed to be used for classification problems with a finite number of classes~\cite{LR}. 
LR utilizes the sigmoid function to map features to a scale of~0 to~1. While training, the 
model derives coefficients for each of the variables and determines a threshold for each 
classification. These coefficients are analogous to the weights in a deep learning model.
While simple, LR models often perform reasonably well on many classification tasks. 

\subsubsection{Overview of Support Vector Machine}

Support Vector Machines (SVM) are a class of popular supervised learning algorithms, 
specifically designed for classification tasks. SVMs have strong generalization capability and 
robustness, and they come in both linear and non-linear forms. 
The SVM input layer accepts the feature vectors, and the prediction is obtained
via the output layer.

The main elements of an SVM are support vectors, decision boundaries (as determined by hyperplanes), 
and a kernel function.  The kernel function can be used to map input data into a higher-dimensional 
``feature space'', which enables the model to deal with non-linear relationships in 
terms of the input data. The main concept behind an SVM is to find the optimal hyperplane 
that can best separate the different classes. Support vectors are those feature vectors
that maximize the margin, where margin is defined as the minimum distance from a feature vector to the 
decision boundary.

The process of training an SVM involves solving a quadratic programming
problem, with the Sequential Minimal Optimization (SMO)
algorithm currently being the best available means to do so. Of relevance to the
research reported in this paper, the SMO algorithm specifically takes advantage of
the fact that the weights of a trained SVM do not require great accuracy~\cite{Stamp_2022}.

\subsubsection{Overview of Multilayer Perceptron}

Multilayer Perceptrons (MLP) are a popular class of feedforward neural network architectures 
that are widely used for supervised learning tasks, including classification and 
regression~\cite{MLP_Intro}. MLPs consist of multiple layers of interconnected nodes, 
where each node receives input from the previous layer and produces output that is passed to 
the next layer.

The input layer of an MLP receives the input data, and the output layer produces the final prediction.
In between these layers, there can be one or more hidden layers that help the model to learn complex patterns 
in data. Each node in the hidden layers applies a nonlinear activation function to the weighted sum of 
its inputs, which helps to capture non-linear relationships in the data.

MLPs are trained using backpropagation, which is an optimization algorithm that adjusts 
the weights of the network based on the difference between the predicted output and the 
actual class label~\cite{Stamp_2022}. The weights are updated using gradient descent, 
which iteratively adjusts the weights to minimize the error.

One of the main advantages of MLPs is their ability to learn complex patterns in the data, making 
them suitable for high-dimensional and non-linear datasets. However, since they use fully-connected
layers, MLPs can be computationally expensive to train
and, as with most DL models, they require a large amount of labeled data to achieve high accuracy.

\subsubsection{Overview of Convolutional Neural Networks}

Convolutional Neural Network (CNN) is a prominent general deep learning technique.
CNNs were originally designed for images, utilizing a unique architecture, consisting of convolutional layers, 
pooling layers, and dense layers (also known as fully-connected layers). The first convolutional layers 
trains filters based on input data. These filters help distinguish basic aspects of the image.
Deeper convolutional layers are trained on the output of the previous layer, which enables
the model to learn more abstract features---and, ultimately, to distinguish between complex images,
such as those representing ``cat'' and ``dog''. Convolution layers are often followed 
by a pooling layer, which decrease the dimensionality, thereby decreasing the computational requirements. 
The final layer of a CNN is a dense layer that is utilized to classify~\cite{Biswal_022}.

In spite of their origin in image classification, CNNs are applicable to other types of data. 
In particular, CNNs can be expected to perform well in cases where local structure is dominant.   

\subsubsection{Overview of LSTM}

Long Short Term Memory (LSTM) is a specific type of Recurrent Neural Network (RNN). 
RNNs allow previous output to be used as input, based on recurrent connections,
which enables such models to have a form of memory that is absent in feedforward architectures. 
However, in plain vanilla RNNs, this memory tends to create gradient flow problems 
when training via backpropagation.
One advantage of LSTMs over plain vanilla RNNs is their ability to mitigate these gradient 
problems when training. LSTMs achieve this improvement over generic RNNs by use of 
a complex gating structure~\cite{Stamp_2022}.

We note in passing that, commercially, LSTM is one of the most successful architectures 
yet developed. Examples of significant applications where LSTMs have played a crucial role
include Google Allo~\cite{allo}, Google Translate~\cite{translate}, 
Apple's Siri~\cite{iBrain}, and Amazon Alexa~\cite{alexa}.

\subsubsection{Overview of VGG16}

Visual Geometry Group 16 (VGG16) is a popular computer vision model~\cite{VGG16}. 
VGG16 was designed as a deep convolutional neural network, pre-trained 
for image classification on the ImageNet dataset.

The model derives its name from its~16 layers with trainable parameters. 
VGG16 includes~13 convolutional layers, five max-pooling layers, and three dense layers, 
resulting in a total of~21 layers. Of these~21 layers, the five 
max-pooling layers do not contain any trainable weights.

One unique aspect of VGG16 is its architectural uniformity. It employs convolutional layers with a 
consistent~$3\times 3$ filter size and a stride of one, using the same padding throughout. 
Additionally, max-pooling layers in VGG16 use a~$2\times 2$ filter with a stride of two. 
This simplicity facilitates ease of implementation and efficient training.

The generalization ability of VGG16 to images beyond its training data has made it a popular
and successful model. VGG16 is commonly employed in transfer learning, where the original 
dense layers are replaced with new task-specific dense layers. The hidden layers, 
consisting of the convolutional and max-pooling layers from the original model, 
remain unchanged and are used as a feature extractor while training the new fully 
connected layers on the new data.

\subsubsection{Overview of DenseNet121}

DenseNet121 is a convolutional neural network architecture that belongs
to the DenseNet family~\cite{dense}. It consists of four dense blocks and several transition 
layers that involve convolution and pooling. The dense layers receive direct input 
from all preceding layers within the same block, allowing for feature reuse. 
Transition layers are inserted between dense blocks to control the spatial dimensions 
and channel depth of the feature maps. A dense blocks is typically followed by an average-pooling 
layer, which serves to reduce the dimensionality. DenseNet121 ends with a classification head, 
containing a fully connected layer with a \texttt{softmax} activation.

DenseNet121 was designed to address the limitations of traditional CNN architectures, 
such as vanishing gradients and information flow constraints. Since its introduction in~2017, 
the model has been successfully applied to image classification tasks and object detection. 
With excellent information flow and feature reuse, DenseNet121 can capture fine-grained 
details and small-scale patterns throughout the network, which is crucial for image analysis. 
In spite of having more than six million trainable parameters,
DenseNet121 is more computationally efficient and requires less memory
than many other comparable CNN models, 
including ResNet152 and VGG16~\cite{dense}.

\subsubsection{Overview of InceptionV3}

InceptionV3 is a prominent CNN architecture that has been very successful 
in the domain of computer vision. This advanced architecture was developed as an 
enhancement to Google's initial Inception model, providing an innovative approach 
to efficient computation and the discernment of complex patterns 
within image data~\cite{inception}.

A distinguishing feature of the InceptionV3 network is its proprietary ``Inception Modules.'' 
These modules incorporate convolution operations with various kernel sizes that operate 
simultaneously, thereby enabling the model to efficiently learn features from the input data.

In typical applications, the input to an InceptionV3 model comprises image data, and its output layer 
delivers predictions across a pre-defined set of classes. The intervening layers of the 
architecture---including numerous convolutional layers, pooling layers, Inception modules, and fully 
connected layers---perform sequential transformations of the input data. This sequence facilitates 
the extraction of patterns and relevant features from the images.

The training of the InceptionV3 model employs backpropagation and gradient descent. Due 
to its complex and deep structure, it also employs advanced techniques such as 
batch normalization (BatchNorm) and sophisticated initialization schemes. 
These approaches are intended to ensure efficient training and 
mitigate potential issues such as the vanishing gradient problem.

The InceptionV3 architecture is known for its balance of computational efficiency and high 
accuracy, performing effectively even with a large number of classes and when handling high-resolution image data. 
Nevertheless, training the InceptionV3 network can be computationally 
intensive, and typically requires a substantial volume of labeled data.

\subsubsection{Overview of Xception}

The Xception model is a deep CNN that is an expansion of the Inception architecture.
The convolutional blocks that make up the Xception architecture each have multiple convolutional layers~\cite{xcept}. 
The convolutions that the Xception model employs are divided into two categories, namely,
depthwise convolutions and pointwise convolutions. Pointwise convolutions 
utilize a~$1\times 1$ convolution to mix the outputs of depthwise convolutions, 
whereas depthwise convolutions apply a single filter to each channel-wise $n\times n$ 
spatial convolution independently. A matrix of pixel values is used to represent the input image, 
and each pixel contains RGB color information, which is passed through an initial convolution block. 
Global average pooling in employed, where the average value of each feature map to create a single value for each 
channel. The final output layer consists of a fully connected layer followed by a \texttt{softmax} activation function 
for classification tasks. 


\subsubsection{Overview of ACGAN}

Auxiliary Classifier Generative Adversarial Network (ACGAN) is a specific type of 
Generative Adversarial Network (GAN) that is used when the data consists of multiple classes. 
In addition to classification, GANs can be used to generate new ``deep fake'' data instances that 
resemble the training data.
					
A GAN has two neural networks, a generator and a discriminator,
that compete in an adversarial zero-sum game. The generator produces new pieces of 
data that are as close to the training data as possible. The discriminator attempts to determine whether 
the input it receives---some of which comes from the generator and some of which comes from the 
actual training data---is generated or authentic. The discriminator and generator weights 
are updated in a way that incentivizes the 
generator to produce ``fake'' data that is similar to the training data, and incentivizes the 
discriminator to accurately diagnose if a sample is fake or real~\cite{Goodfellow2014_long}.
					
An ACGAN works similarly, except that the discriminator also returns the class it 
thinks the data belongs to. The ACGAN incentivizes the generator to produce believable 
fakes that conform well to a specific class, while the discriminator is incentivized to accurately 
diagnose fake samples and classify the data. 
 
\subsection{Related Work}

In the paper~\cite{EvilModel}, a technique that the authors refer to as 
``EvilModel'' is used to hide malware in a neural network model. In one example, 
a malware sample of size~36.9~MB is embedded in a specific model, 
and the accuracy of the model is reduced by about~1\%. 
The authors of~\cite{EvilModel} embed malware
in a learning model by carefully selecting 
weights that have minimal effect 
on model performance, and then overwrite these
weights with the malware sample.

The paper~\cite{EvilModel2} is a continuation of the work in~\cite{EvilModel}. Among other results,
in~\cite{EvilModel2}, malware is embedded in the least significant
bits of model weights, and an ``embedding rate'' of slightly more
than~48\%\ is achieved.

The paper~\cite{StegoNet} considers a technique that its authors call ``StegoNet.''
Among other contributions, this paper includes experiments consisting of modifying 
the least significant bits of model weights, and they propose 
a plausible trigger mechanisms for malware that is embedded in
a machine learning model.

Here, we consider the problem of embedding information in the least significant bits 
of model weights. In comparison to~\cite{EvilModel2}, 
we are generally able to achieve relatively high embedding rates 
with no significant decrease in model performance.
In contrast to both~\cite{EvilModel2} and~\cite{StegoNet}, we
consider far more model types, and our analysis is much more
thorough, as we provide graphs explicitly showing the 
tradeoff between the number of bits overwritten and model accuracy.

The work presented in this paper is a continuation of the work in~\cite{Lei},
where the steganographic capacity of a Multilayer Perceptron (MLP), 
a Convolutional Neural Network (CNN), and a specific Transformer model
are analyzed. Here, we consider the models introduced 
in Section~\ref{sect:learningModels}, above, 
and provide a detailed analysis of the steganographic capacity of each.

\section{Implementation}\label{chap:3}

In this section, we introduce the malware dataset used to train our learning models. 
Then we provide details on our experimental design.
Our experimental results are given in Section~\ref{chap:4}, below.

\subsection{Dataset}

Malware samples that are closely related can be grouped into families. 
Malware samples within a family generally have similar functionality, behavior, and code structure. 
Members of a given family typically share a core code base 
that contains common functions, routines, and behavior. Malware families
tend to evolve over time, and new families can branch off from existing families.

In this research, we consider a malware dataset obtained from VirusShare~\cite{VirusShare}. 
This dataset contains more than~500,000 malware executables.
From this dataset of~500,000 malware executables, we consider the
top~10 most numerous families---these malware families and number of 
samples per family are listed Table~\ref{tab:1}. Note that the dataset is imbalanced, 
with the most numerous of the~10 families containing more than~17\%\ of the
samples, while the least numerous has slightly over~7\%\ of the samples.


\begin{table}[!htb]
\caption{Malware families}\label{tab:1}
\centering
\adjustbox{scale=0.85}{
\begin{tabular}{c|cc}\midrule\midrule
Family & Samples & Fraction of total\\ \midrule
\texttt{Adload} & 1225 & 0.0798 \\
\texttt{BHO} & 1412 & 0.0920 \\
\texttt{Ceeinject} & 1084 & 0.0706 \\
\texttt{OnLineGames} & 1511 & 0.0984 \\
\texttt{Renos} & 1567 & 0.1020 \\
\texttt{Startpage} & 1347 & 0.0877 \\
\texttt{VB} & 1110 & 0.0723 \\
\texttt{VBinject} & 2689 & 0.1751 \\
\texttt{Vobfus} & 1108  & 0.0721 \\
\texttt{Winwebsec} & 2303  & 0.1500 \\
\midrule
Total & 15,356 & 1.0000 \\ \midrule\midrule
\end{tabular}
}
\end{table}


Next, we briefly describe each of these families; for more details,
see~\cite{Lei}. These families include several different categories
of malware, including viruses, worms, and Trojans.

\begin{description}

\item[\texttt{Adload}\!] is an adware program that displays unwanted advertisements 
on a web browser~\cite{Adload}. 

\item[\texttt{BHO}\textrm{,}\!] is a type of add-on or 
plugin for web browsers, such as Internet Explorer. While there are many legitimate BHOs, the 
malware version can perform unwanted actions, such as redirecting web traffic or displaying 
unwanted ads~\cite{BHO}. 

\item[\texttt{Ceeinject}\!] injects itself into legitimate processes running on a Windows operating system, 
allowing it to execute its malicious code undetected~\cite{Ceeinject}.

\item[\texttt{OnLineGames}\!] is a Trojan that mimics an online game~\cite{OnLineGames}.

\item[\texttt{Renos}\!] is designed to trick users into purchasing 
fraudulent security software or services~\cite{Renos}. 

\item[\texttt{Startpage}\!] is a family is Trojans that modifies a user's web browser settings, such as 
the homepage and search engine, without the user's consent~\cite{Startpage}. 

\item[\texttt{VB}\!] is a simple Trojan that spreads a worm by copying 
itself to removable drives, network shares, and other accessible file systems~\cite{VB}. 

\item[\texttt{VBinject}\!] is a
general technique that is applied by malware author to inject malicious program into legitimate 
Windows processes~\cite{VBinject}. 

\item[\texttt{Vobfus}\!] is a malware family that downloads other malware, such as \texttt{Zbot},
onto a victim’s computer~\cite{Vobfus}. 

\item[\texttt{Winwebsec}\!] is designed to trick users into purchasing fraudulent security software or 
services. It displays false alerts and warnings about supposed security threats~\cite{Winwebsec}.  

\end{description}

We consider several types of feature vectors, depending on the requirements
of the particular model under consideration. For our feedforward models
(LR, SVM, MLP, ACGAN), we
extract a relative byte histogram from each sample. For our image-based
models (CNN, VGG16, DenseNet121, InceptionV3, Xception), 
we treat the raw bytes of an \texttt{exe} file as an image.
For example, if a model uses~$64\times 64$ images, we place the first~4096
bytes of an \texttt{exe} into a~$64\times 64$ array (padding with~0 bytes, 
if necessary) which we then treat as an image. For our model that requires
sequential data (LSTM), we use the first~$n$ bytes of each \texttt{exe} file.
Note that in all cases, these feature vectors are trivial to generate, and require no costly
disassembly or dynamic analysis.

\subsection{Model Training}

A similar training and testing procedure is used for each of the~10 learning models considered. 
First, we train each model with labeled data and test the trained model, which establishes 
a baseline level of performance. In this phase a grid search is performed over a set of reasonable
hyperparameter values. Accuracy is used as our measure of performance.
 
After the initial training and testing, data is inserted into the low-order~$n$ bits of the weights, 
which, on average, changes about half of the bit values. For each~$n$, the performance of the 
model is re-evaluated using the same data and accuracy metric as for the unmodified model.
This allows for a direct comparison of the results for each~$n$. We graph the accuracy
as a function of~$n$.

\section{Steganographic Capacity Experiments}\label{chap:4}

In this section, we consider the steganographic capacity of each of the~10 models discussed in
Section~\ref{sect:learningModels}. As mentioned above, to measure the steganographic capacity, 
we embed information in the low-order~$n$ bits of the model weights, and
we graph the accuracy as a function of~$n$. 
In all cases, the information that we hide is extracted from the 
\texttt{pdf} version of the book \textit{Alice's Adventures in Wonderland}~\cite{Alice}.

For each deep learning model, we consider the following cases.
\begin{enumerate}
\item Only the output layer weights are modified
\item The weights of all hidden layers are modified
\item All of the model weights are modified
\end{enumerate}
For selected models, we also consider the effect of overwriting
the weights of individual layers.
In addition to graphing the model accuracy as a function of~$n$, 
we provide a capacity graph, that is, the number of model bits that 
have been overwritten for each~$n$. 

To determine the overall capacity of a model, we find the number of bits~$n$
that must be overwritten for a~1\%\ drop in accuracy, as compared to the original trained model,
which has no bits of its weights overwritten. We then use~$n-1$ as the per-weight
steganographic capacity, and the total capacity (in bits) is determined by multiplying
the number of weights by~$n-1$. We give the capacity in kilobytes (KB) or
megabytes (MB), as appropriate.



\subsection{LR Experiments}

This model utilized the \texttt{LogisticRegression()} class from the \texttt{sklearn} 
package in Python \texttt{scikit-learn}. The class has~4 different hyperparameters, 
all of which were tested via grid search and optimized. The hyperparameter values
tested are given in Table~\ref{tab:LR}, with the values in boldface yielding the best results.

\begin{table}[!htb]
\caption{LR hyperparameters tested}\label{tab:LR}
\centering
\adjustbox{scale=0.85}{
\begin{tabular}{c|c}\midrule\midrule
Hyperparameter & Values tested\\ \midrule
\texttt{solver} & \textbf{lbfgs}, saga, liblinear \\
\texttt{penalty} & elasticnet, \textbf{l2} \\
\texttt{C} (regularization) & 0.2, 0.3, 0.5, 0.7, \textbf{0.8} \\
\texttt{max\_iter} & 50, 80, \textbf{100}, 120, 200, 500 \\ \midrule\midrule
\end{tabular}
}
\end{table}

For this 10-class classification problem, the model achieves a respectable accuracy of~0.8717 on the validation set. 
From the confusion matrix for our model, which appears in Figure~\ref{fig:confLR} in the appendix, 
we see a similar spread of errors, as compared to the other models tested, with slightly poorer 
performance in identifying \texttt{WinWebSec} viruses. 

\begin{figure}[!htb]
\centering
\adjustbox{scale=0.75}{
\input{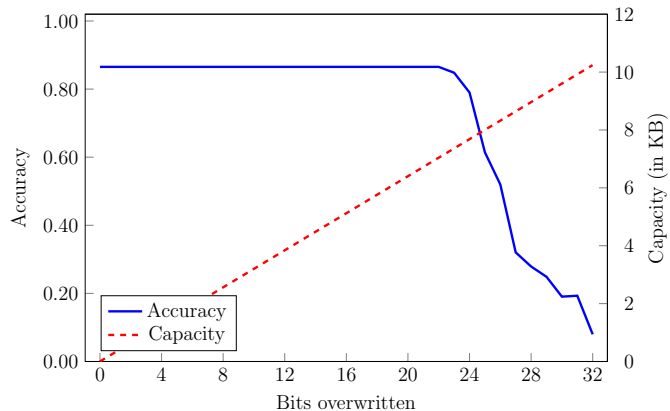}
}
\caption{LR steganographic capacity graph}\label{fig:LRoverwrite}
\end{figure}

Since LR models only have one layer of coefficients, we can only overwrite the bits in that layer;
the graph of these results are given in Figure~\ref{fig:LRoverwrite}. There is no drop in model accuracy
when~$n\leq 22$ bits are overwritten, with about a~2\%\ drop at~$n=23$. Therefore, we can overwrite
the~22 low-order bits of each weight with no loss in performance and hence we deem~$n=22$ as the 
steganographic capacity per weight of this model. Since the model has~2560 weights, the total steganographic
capacity is~$22\cdot 2560 = 56{,}320$ bits, or~7.04~KB.

\subsection{SVM Experiments}

The {\tt svm.SVC()} function from the \texttt{sklearn} module from \texttt{scikit-learn} 
was used for the training and testing of our SVM model. The hyperparameters that we 
tested are listed in Table~\ref{tab:SVM}, with the selected values in boldface. 
For example, the SVM model with a \texttt{C} value of~1 yielded the best results,
and the \texttt{linear} kernel was selected, with a \texttt{gamma} value of~0.1.

\begin{table}[!htb]
\caption{SVM model hyperparameters tested}\label{tab:SVM}
\centering
\adjustbox{scale=0.85}{
\begin{tabular}{c|c}\midrule\midrule
Hyperparameter & Values tested\\ \midrule
\texttt{C} (regularization) & 0.1, \textbf{1}, 10 \\
\texttt{kernel} & \textbf{linear}, rbf  \\
\texttt{gamma} & \textbf{0.1}, 1, 10 \\\midrule\midrule
\end{tabular}
}
\end{table}

Based on the confusion matrix for the SVM model, which appears in Figure~\ref{fig:confSVM} in the appendix, 
the model performs similarly to the other models, with the highest level of confusion for the \texttt{VBinject} 
class of viruses. Also, the SVM model outputs a classifying accuracy of~0.8264 
for the \texttt{Adload} class, which is lower than the most accurate of our models.

\begin{figure}[!htb]
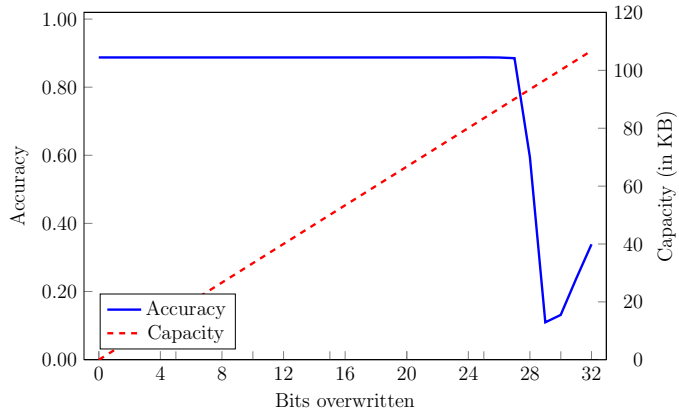

\centering
\adjustbox{scale=0.75}{
\input figures/svm_graph.tex
}
\caption{SVM  steganographic capacity graph}\label{fig:graph}
\end{figure}

The overall accuracy of our SVM model is~0.8870. An SVM consists of a single ``layer'' 
of coefficients, which correspond to the weights of a deep learning model.
Furthermore, SVM coefficients are within the range of~$\hbox{}-1$ to~1, with a higher magnitude 
indicating a larger importance in determining the decision boundary.
The model was able to withstand the overwriting of~27 bits before 
experiencing a significant drop in accuracy, 
which is a slightly higher per-weight capacity than any of the deep learning models considered.
The SVM model contains~26,703 coefficients (i.e., weights) and hence we calculate 
the steganographic capacity of the model to be~90.12~KB.

\subsection{MLP Experiments}

The MLP results we present here are from~\cite{Lei}; we include these results for 
the sake of comparison.
The {\tt MLPClassifier()} from the \texttt{sklearn.neural\_network} module was used to 
train and test our MLP model. The hyperparameters tested are listed in Table~\ref{tab:3}, 
with the selected values appear in boldface. Note that a model with two hidden layers, 
with~128 and~10 neurons, respectively, was best. Also, the \texttt{logistic} function was 
selected as our activation function.   

\begin{table}[!htb]
\caption{MLP model hyperparameters tested}\label{tab:3}
\centering
\adjustbox{scale=0.85}{
\begin{tabular}{c|c}\midrule\midrule
Hyperparameter & Values tested\\ \midrule
\texttt{hidden\_layer\_sizes} & (64, 10), (96, 10), \textbf{(128, 10)} \\
\texttt{activation} & identity, \texttt{\textbf{logistic}}  \\
\texttt{alpha} & 0.0001, \textbf{0.05} \\
\texttt{random\_state} & 30, \textbf{40}, 50 \\ 
\texttt{solver} & \texttt{\textbf{adam}} \\
\texttt{learning\_rate\_init} &  \textbf{0.00001} \\
\texttt{max\_iter} &  \textbf{10000} \\\midrule\midrule
\end{tabular}
}
\end{table}

\begin{figure}[!htb]
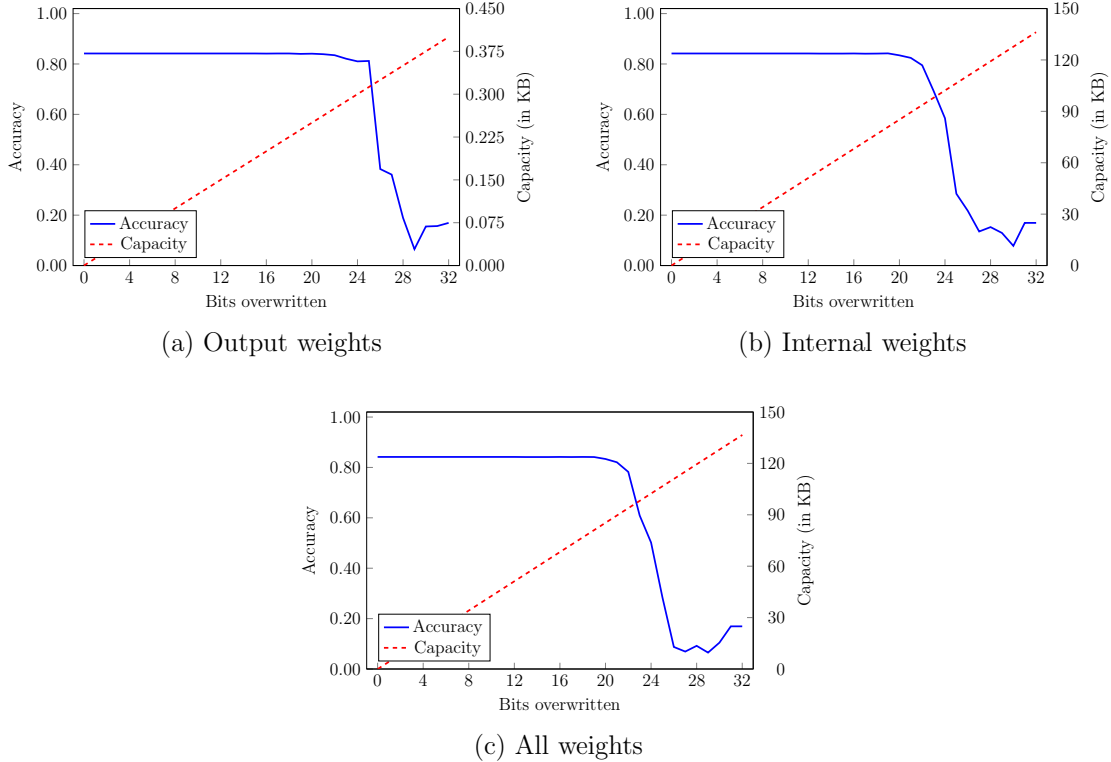

\centering
\begin{tabular}{cc}
\adjustbox{scale=0.925}{
\input figures/mlp_output.tex
}
&
\adjustbox{scale=0.925}{
\input figures/mlp_internal.tex
}
\\
\adjustbox{scale=0.85}{(a) Output weights}
&
\adjustbox{scale=0.85}{(b) Internal weights}
\\ \\
\multicolumn{2}{c}{\adjustbox{scale=0.925}{
\input figures/mlp_all.tex
}}
\\
\multicolumn{2}{c}{\adjustbox{scale=0.85}{(c) All weights}}
\end{tabular}
\caption{MLP steganographic capacity graphs~\cite{Lei}}\label{fig:MLP_Plot}
\end{figure}

The results obtained when overwriting the low order bits of all weights 
of our trained MLP model are summarized in Figure~\ref{fig:MLP_Plot}(c). We observe that
the original accuracy for the model is~0.8417, and the performance of the model 
is unchanged when the low-order~19 bits of the weights
are overwritten, while there is a~1\%\ drop in performance when~20 bits are overwritten. 
Overwriting more bits causes the accuracy to drop substantially.

Figures~\ref{fig:MLP_Plot}(a) and~(b) are the results when overwriting 
the output and internal layer weights, respectively. The results in these two cases are similar---although
not identical---to the results for all weights, discussed above.

There are~100 weights in the output layer, and~34,048 weights in the hidden layer, 
which makes the total number of weights~34,148 in this particular MLP model. 
Since we can hide information in~19 bits of the all of the weights, we find that the
steganographic capacity of this MLP model is approximately~81.10~KB.

\subsection{CNN Experiments}

A Keras \texttt{Sequential} model with the {\tt Conv2D()}, {\tt Dense()}, and {\tt MaxPooling2D()} 
layers provided by \texttt{tensorflow.keras.layers} was used to train our CNN model. 
After testing the hyperparameters listed in Table~\ref{tab:CNN}, we found those in boldface 
to be the optimal choices for our model. The~12 layers in our CNN consisted of four {\tt Conv2D()} 
and {\tt MaxPooling2D()} layers, along with two {\tt Dense()} layers. The other two layers are 
dropout and flattening layers, for which the placement and dropout rate were tested. 
The activation function for the last dense layer is \texttt{softmax}, with the other convolution layers 
using \texttt{ReLU} as their activation functions.  

\begin{table}[!htb]
\caption{CNN hyperparameters tested}\label{tab:CNN}
\centering
\adjustbox{scale=0.85}{
\begin{tabular}{c|c}\midrule\midrule
Hyperparameter & Values tested\\ \midrule
\texttt{layers} & 6, 8, 10 \textbf{12} \\
\texttt{activation} & \texttt{ReLU}, \textbf{softmax}, \texttt{sigmoid} \\
\texttt{dropout rate} & 0.1, \textbf{0.2}, 0.3, 0.4, 0.5 \\
\texttt{learning rate} & \textbf{0.001}, 0.1 \\ \midrule\midrule
\end{tabular}
}
\end{table}

\begin{figure}[!htb]
\centering
\adjustbox{scale=0.65}{
\input{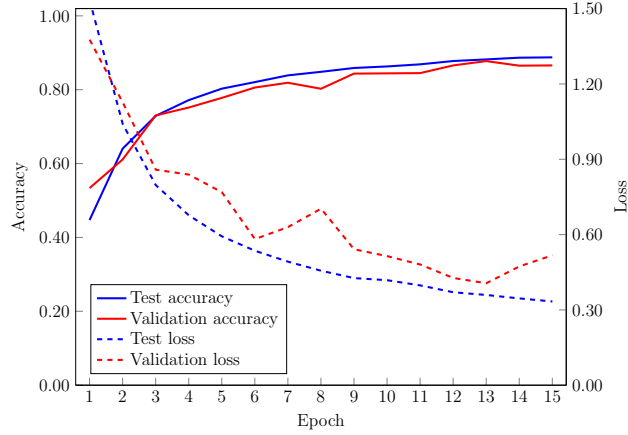}
}
\caption{Test and validation graphs for CNN}\label{fig:CNNtestvalid}
\end{figure}

Our CNN model achieves an accuracy of~0.8925. 
From the accuracy and loss graph in Figure~\ref{fig:CNNtestvalid},
we detect no signs that the model is overfitting the data.

The confusion matrix for our best CNN model appears in
Figure~\ref{fig:confCNN} in the appendix.
From the confusion matrix, we observe that the \texttt{VB} and \texttt{VBInject} viruses
account for the majority of errors on the test set. This is reasonable, as these two families
are relatively similar.

\begin{figure}[!htb]
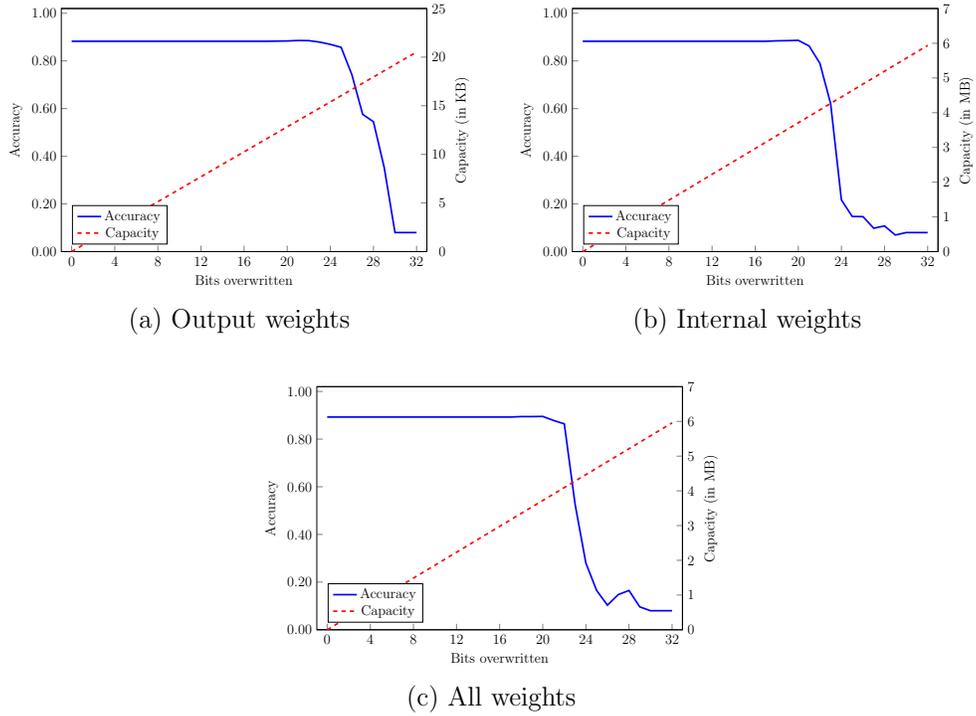

\centering
\begin{tabular}{cc}
\adjustbox{scale=0.525}{
\input figures/cnn_output.tex
}
&
\adjustbox{scale=0.525}{
\input figures/cnn_internal.tex
}
\\
\adjustbox{scale=0.85}{(a) Output weights}
&
\adjustbox{scale=0.85}{(b) Internal weights}
\\ \\
\multicolumn{2}{c}{\adjustbox{scale=0.525}{
\input figures/cnn_all.tex
}}
\\
\multicolumn{2}{c}{\adjustbox{scale=0.85}{(c) All weights}}
\end{tabular}
\caption{CNN steganographic capacity graphs}\label{fig:CNN_Plot}
\end{figure}

The results of overwriting the low-order bits for different layers can be seen in 
Figure~\ref{fig:CNN_Plot}. In the case of all model weights, the accuracy first drops when
we overwrite~21 bits, and hence we denote the per-weight capacity as~20 bits.
Our CNN model has~5130 weights in the output layer and~1,484,544 weights in the internal layers, 
for a total of~1,489,674 weights. This give us an overall steganographic capacity of
approximately~3.72~MB

\subsection{LSTM Experiments}

The {\tt LSTM()} function from the \texttt{keras} module was used for training and testing our LSTM model. 
Table~\ref{tab:lstm} shows the hyperparameters tested while training, and the boldface entries indicate the 
combination that yielded the best results.  The confusion matrix for our best LSTM model appears in
Figure~\ref{fig:confLSTM} in the appendix.

\begin{table}[!htb]
\caption{LSTM model hyperparameters tested}\label{tab:lstm}
\centering
\adjustbox{scale=0.85}{
\begin{tabular}{c|c}\midrule\midrule
Hyperparameter & Values tested\\ \midrule
\texttt{batch\_size} & 16, \textbf{32}, 64, 128 \\
\texttt{activation} & \textbf{tanh}, \texttt{ReLU}  \\
\texttt{epoch} & 5, \textbf{10}, 12 \\ 
\texttt{optimizer} & \texttt{\textbf{RMSprop}, adam} \\
\texttt{learning\_rate} &  0.0001, \textbf{0.001} \\
\texttt{LSTM\_units} &  64, 128, \textbf{512} \\
\texttt{dense\_layer\_units} &  64, \textbf{128}, \\
\texttt{sequence\_length} & 150, 200, \textbf{300}, 350, 400 \\ \midrule\midrule
\end{tabular}
}
\end{table}

As feature vectors for our LSTM, we use the first~$N$ bytes of the \texttt{exe} files, 
where each byte is converted to the range of~0 and~1 by treating the byte value as an integer
and dividing by~255. We experimented with the different values of~$N$ as listed in Table~\ref{tab:lstm}
and found that~$N=300$ gave us the best results. Note that this model is extremely lightweight, 
and hence it is not surprising that it yields slightly less accurate results, as compared
to other models tested. 

\begin{figure}[!htb]
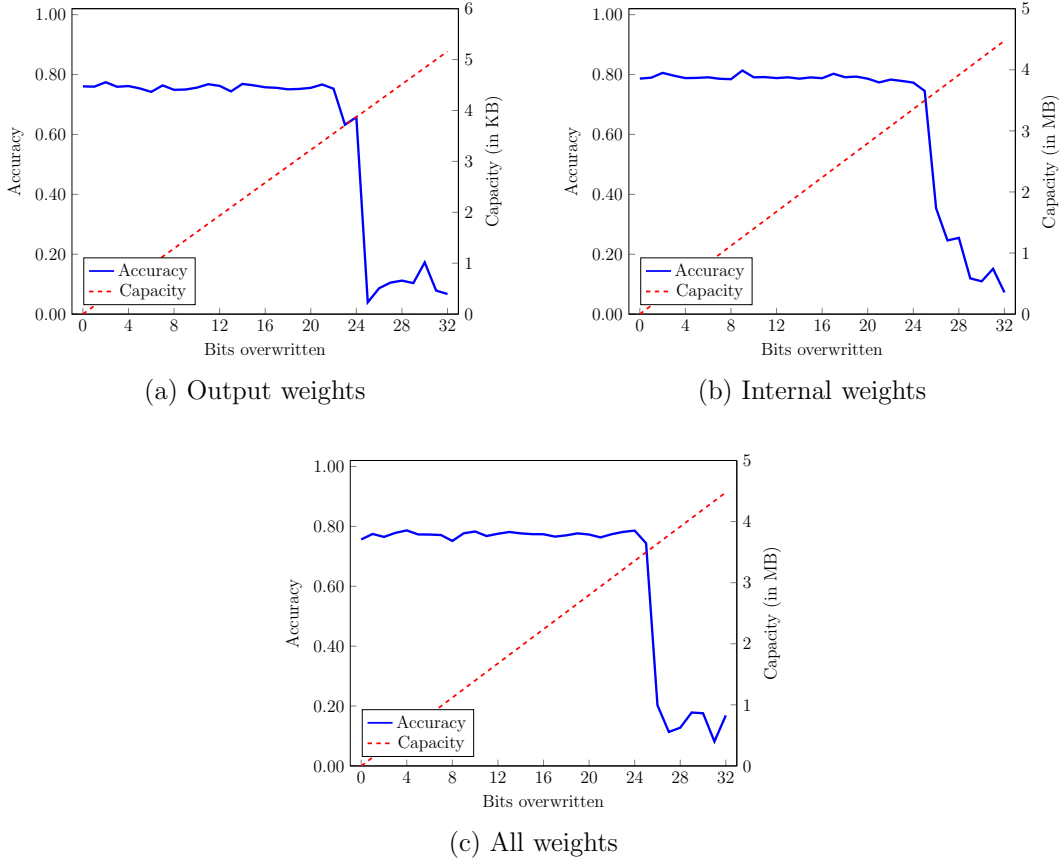

\centering
\begin{tabular}{cc}
\adjustbox{scale=0.925}{
\input figures/lstm_output.tex
}
&
\adjustbox{scale=0.925}{
\input figures/lstm_internal.tex
}
\\
\adjustbox{scale=0.85}{(a) Output weights}
&
\adjustbox{scale=0.85}{(b) Internal weights}
\\ \\
\multicolumn{2}{c}{\adjustbox{scale=0.925}{
\input figures/lstm_all.tex
}}
\\
\multicolumn{2}{c}{\adjustbox{scale=0.85}{(c) All weights}}
\end{tabular}
\caption{LSTM steganographic capacity graphs}\label{fig:LSTM_Plot}
\end{figure}

When overwriting low-order bits of all weights, the validation accuracy is slightly more than~0.78 up to~24 bits. 
However, the accuracy drops about~4\%\ when~25 bits have been modified per weight, before plummeting 
at~26 bits, as shown in Figure~\ref{fig:LSTM_Plot}(c). The results for the output and internal layers
are similar.

With~1,119,626 trainable parameters, the weights were split into 
an LSTM layer and two dense layers. 
The majority of the units are found in the LSTM layer (containing a total of~1,052,672) 
in this particular LSTM model. The first dense layer has~65,664 weights while the output dense 
layer only possess~1290. Based on overwriting the~24 low-order bits, the total steganographic 
capacity of this LSTM is about~3.36~MB. 

\subsection{VGG16 Experiments}\label{sect:VGG}

The {\tt VGG16()} model pre-trained on ImageNet from the \texttt{tf.keras.applications} module was used to 
train and test our VGG16 model. Since this is a transfer learning model, we replaced the old dense layers 
with a new dense layer that has~10 units, each unit corresponding to one of our output classes. The output layer uses 
a \texttt{softmax} activation function. The hyperparameters tested are listed in Table~\ref{tab:vgg}, 
with the selected values in boldface. Note that most of the hyperparameters of the model are predetermined 
due to transfer learning. The confusion matrix for our best VGG16 model appears in
Figure~\ref{fig:confVGG} in the appendix.

\begin{table}[!htb]
\caption{VGG16 model hyperparameters tested}\label{tab:vgg}
\centering
\adjustbox{scale=0.85}{
\begin{tabular}{c|c}\midrule\midrule
Hyperparameter & Values tested\\ \midrule
\texttt{random\_state} & 100, \textbf{120}, 130 \\ 
\texttt{solver} & \texttt{\textbf{adam}} \\
\texttt{learning\_rate\_init} & 0.001, \textbf{0.01} \\
\texttt{max\_iter} &  50, 75, \textbf{100} \\\midrule\midrule
\end{tabular}
}
\end{table}

For all of the pre-trained models considered here 
(i.e., VGG16, DenseNet121, InceptionV3, and Xception)
we refer to the weights that are re-trained for our
malware classification problem as the ``trained weights.''
These are in contrast to the pre-trained weights, 
which do not change from the pre-trained models.

\begin{figure}[!htb]
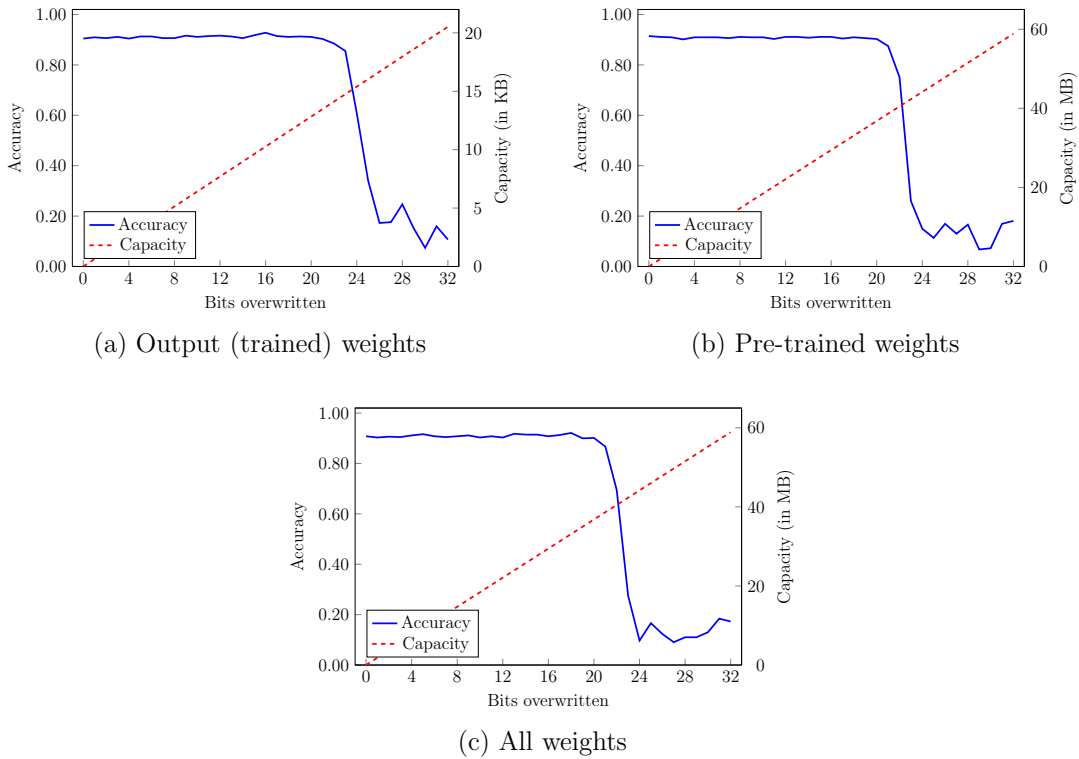

\centering
\begin{tabular}{cc}
\adjustbox{scale=0.925}{
\input figures/vgg16_out.tex
}
&
\adjustbox{scale=0.925}{
\input figures/vgg16_in.tex
}
\\
\adjustbox{scale=0.85}{(a) Output (trained) weights}
&
\adjustbox{scale=0.85}{(b) Pre-trained weights}
\\ \\
\multicolumn{2}{c}{\adjustbox{scale=0.925}{
\input figures/vgg16_all.tex
}}
\\
\multicolumn{2}{c}{\adjustbox{scale=0.85}{(c) All weights}}
\end{tabular}
\caption{VGG16 steganographic capacity graphs}\label{fig:VGG_plot}
\end{figure}

Only the output layer weights of this model were retrained for
our malware classification problem. 
From the graph in Figure~\ref{fig:VGG_plot}(a), we see that our VGG16 model
accuracy is maintained when~21 bits of the trained weights
are overwritten, with a drop of more than~2\%\ at~22 bits, and a larger drop thereafter.
Thus, the per-weight capacity is~21 bits when only trained weights are considered.
Since the output layer has~5130 weights, this gives us a capacity of~13.47~KB.

Figure~\ref{fig:VGG_plot}(b) gives capacity results for the pre-trained weights,
while Figure~\ref{fig:VGG_plot}(c) contains the results for all weights.
In both of these cases, the per-weight capacity is~20 bits.
The hidden layer of our VGG16 implementation has~14,714,688 weights, 
and hence the total number of weights is~14,719,818 in our VGG16 model. 
considering all weights, this gives us a capacity of almost~36.8~MB. 

Figure~\ref{fig:VGG_layers} in the appendix gives the steganographic
capacity results for each of the~13 individual layers in our VGG16 model. In each case, 
these graphs follow a similar pattern, and hence we observe no dramatic differences between 
the layers, with respect to our steganographic capacity experiments.

\subsection{DenseNet121}

{\tt DenseNet121()} from the \texttt{tensorflow} module was used for training and testing. 
Table~\ref{tab:dense} shows the hyperparameters tested, and the boldface entries indicate 
the combination that attained the best results. 
The confusion matrix for our best DenseNet121 model appears in
Figure~\ref{fig:confDense} in the appendix.

\begin{table}[!htb]
\caption{DenseNet121 hyperparameters tested}\label{tab:dense}
\centering
\adjustbox{scale=0.85}{
\begin{tabular}{c|c}\midrule\midrule
Hyperparameter & Values tested\\ \midrule
\texttt{batch\_size} & 16, 32, \textbf{64}, 128 \\
\texttt{activation} & \texttt{\textbf{ReLU}}  \\
\texttt{kernel\_regularizer} & \textbf{l2 (0.01)} \\
\texttt{epoch} & 5, \textbf{10}, 12 \\ 
\texttt{optimizer} & \texttt{\textbf{adam}} \\
\texttt{learning\_rate} &  \textbf{0.0001}, 0.001 \\
\texttt{dense\_layer\_units} &  64, 128, \textbf{512} \\
\texttt{input\_shape} & 64, 128, 224, \textbf{64} \\ \midrule\midrule
\end{tabular}
}
\end{table}

\begin{figure}[!htb]
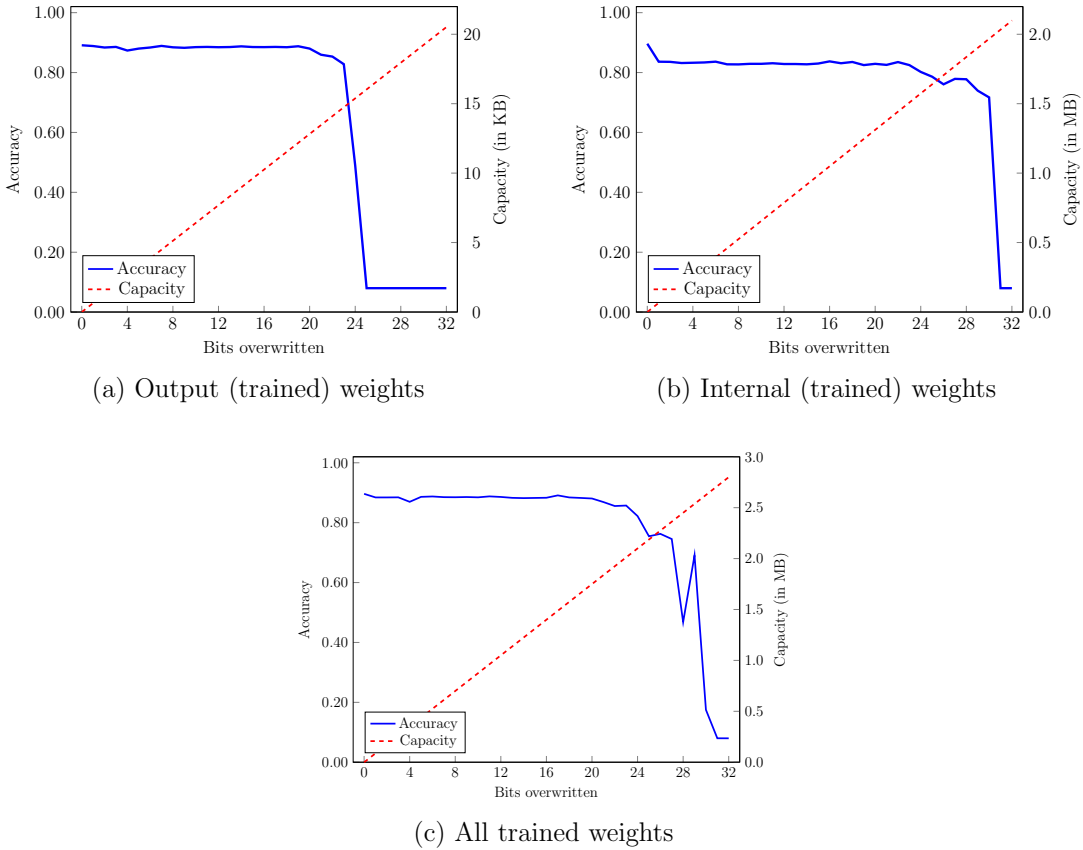

\centering
\begin{tabular}{cc}
\adjustbox{scale=0.925}{
\input figures/densenet_output.tex
}
&
\adjustbox{scale=0.925}{
\input figures/densenet_internal.tex
}
\\
\adjustbox{scale=0.85}{(a) Output (trained) weights}
&
\adjustbox{scale=0.85}{(b) Internal (trained) weights}
\\ \\
\multicolumn{2}{c}{\adjustbox{scale=0.925}{
\input figures/densenet_all.tex
}}
\\
\multicolumn{2}{c}{\adjustbox{scale=0.85}{(c) All trained weights}}
\end{tabular}
\caption{DenseNet121 steganographic capacity graphs}\label{fig:Dense_Plot}
\end{figure}

From Figure~\ref{fig:Dense_Plot}(c) we observe that the model accuracy is about~0.88
and that overwriting~20 bits of the trained weights provides
no loss in accuracy, but overwriting~21 bits results in a~2\%\ drop,
with further declines thereafter. Thus, the per-weight steganographic capacity
of our DenseNet121 model is~20, when considering the trained weights.

DenseNet121 contains~7,571,530 total parameters, but only~700,106 weights are trainable.
Thus, when modifying the trained weights, the model has a capacity of about~1.75~MB, 
based on a per-weight capacity of~20 bits. 

\subsection{InceptionV3}\label{sect:V3}

The InceptionV3 pre-trained model from the \texttt{Keras} library was utilized for our training
and testing. This model is based on transfer learning, with fine tuning applied to the final layers 
(i.e., the output and dense layers)
for our specific malware classification problem. The hyperparameters tested are listed in Table~\ref{tab:incept}, 
with the selected values in boldface. Since InceptionV3 is a pre-trained model, 
only three hyperparameters tested, namely, \texttt{epochs}, \texttt{batch\_size}, and \texttt{learning\_rate}.
The confusion matrix for our best InceptionV3 model appears in
Figure~\ref{fig:confIncept} in the appendix. 

\begin{table}[!htb]
\caption{InceptionV3 hyperparameters tested}\label{tab:incept}
\centering
\adjustbox{scale=0.85}{
\begin{tabular}{c|c}\midrule\midrule
Hyperparameter & Values tested\\ \midrule
\texttt{epochs} & 2, \textbf{4}, 5, 8 \\
\texttt{batch\_size} & 32, \textbf{64}, 128 \\
\texttt{learning\_rate} & 0.001, \textbf{0.0001}, 0.0005 \\ \midrule\midrule
\end{tabular}
}
\end{table}

\begin{figure}[!htb]
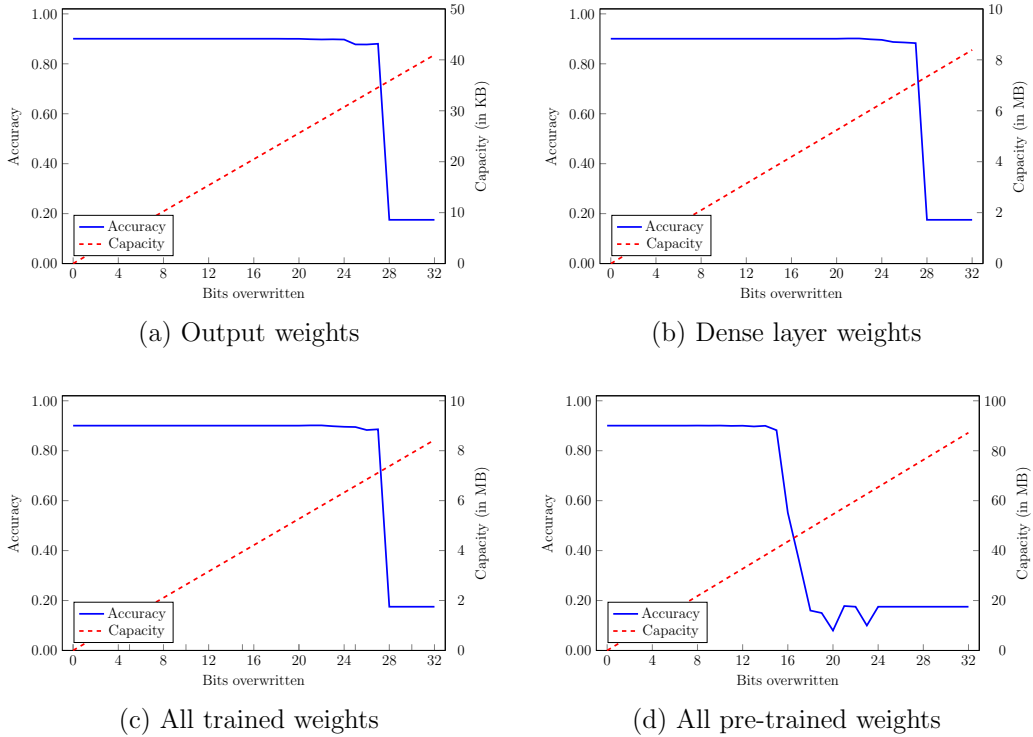

\centering
\begin{tabular}{cc}
\adjustbox{scale=0.55}{
\input figures/inceptionv3_output.tex
}
&
\adjustbox{scale=0.55}{
\input figures/inceptionv3_dense.tex
}
\\
\adjustbox{scale=0.85}{(a) Output weights}
&
\adjustbox{scale=0.85}{(b) Dense layer weights}
\\ \\
\adjustbox{scale=0.55}{
\input figures/inceptionv3_train.tex
}
&
\adjustbox{scale=0.55}{
\input figures/inceptionv3_inceptionV3layer.tex
}
\\
\adjustbox{scale=0.85}{(c) All trained weights}
& 
\adjustbox{scale=0.85}{(d) All pre-trained weights}
\end{tabular}
\caption{InceptionV3 steganographic capacity graphs}\label{fig:InceptionV3_Plot}
\end{figure}

Figure~\ref{fig:InceptionV3_Plot}(c) summarize the effect of hiding data in all trained weights of
our InceptionV3 model. The model's initial accuracy is approximately~0.9004 and remains above~0.89
until we have overwritten the~26 least-significant bits, which causes only a slight decline in accuracy to~0.88,
with more substantial drops thereafter. Thus, with respect to the trained weights,
we consider~25 bits as the per-weight capacity of this model.

Our InceptionV3 model has~10,240 weights in the output layer, and~2,097,152 weights in the dense layer,
for a total of~2,107,392 trained weights. With a per-weight capacity of~25 bits,
this gives us a total steganographic capacity of approximately~6.59~MB in the trained weights.

In Figure~\ref{fig:InceptionV3_Plot}(d), we have the capacity graph for the pre-trained
weights of the InceptionV3 model. Interestingly, the pre-trained weights have a much lower per-weight
steganographic capacity, as compared to the trained weights. For the 
pre-trained weights, we observe a drop of about~1\%\ in accuracy at~15 bits,
followed by a steep drop at~16 bits, and hence we consider~14 bits as the
per-weight capacity with respect to the pre-trained weights. There
are~21,802,784 weights in the pre-trained InceptionV3 layer, so even with 
its lower per-weight capacity of~14 bits, the total steganographic capacity of the
pre-trained weights is large, at~38.15~MB.

\subsection{Xception}

The {\tt Xception} pre-trained model from the \texttt{tensorflow.keras} module was used for
our Xception experiments. The hyperparameters tested are listed in Table~\ref{tab:xcept}, 
and the combination that yielded the best result appear in boldface. Note that both \texttt{softmax} and 
\texttt{ReLU} 
activation functions were utilized in the hidden layers, and the input data was reshaped to fit the 
input size of~$(256,256,3)$.
The confusion matrix for our best Xception model appears in Figure~\ref{fig:confIncept} 
in the appendix.

\begin{table}[!htb]
\caption{Xception model hyperparameters tested}\label{tab:xcept}
\centering
\adjustbox{scale=0.85}{
\begin{tabular}{c|c}\midrule\midrule
Hyperparameter & Values tested\\ \midrule
\texttt{input\_shape} & \textbf{(256,256,3)}, (299,299,3) \\
\texttt{activation} & \texttt{ReLU}, \texttt{softmax}  \\
\texttt{num\_classes} & \textbf{10} \\
\texttt{batch\_size} & 16, 32, \textbf{64} \\ 
\texttt{epochs} & 5, 7, \textbf{10}, 15 \\
\texttt{learning\_rate} &  0.001, \textbf{0.0001}\\
\texttt{kernal\_regularizer} & \textbf{l2(0.01)} \\
\texttt{test\_split} &  \textbf{0.2} \\\midrule\midrule
\end{tabular}
}
\end{table}

\begin{figure}[!htb]
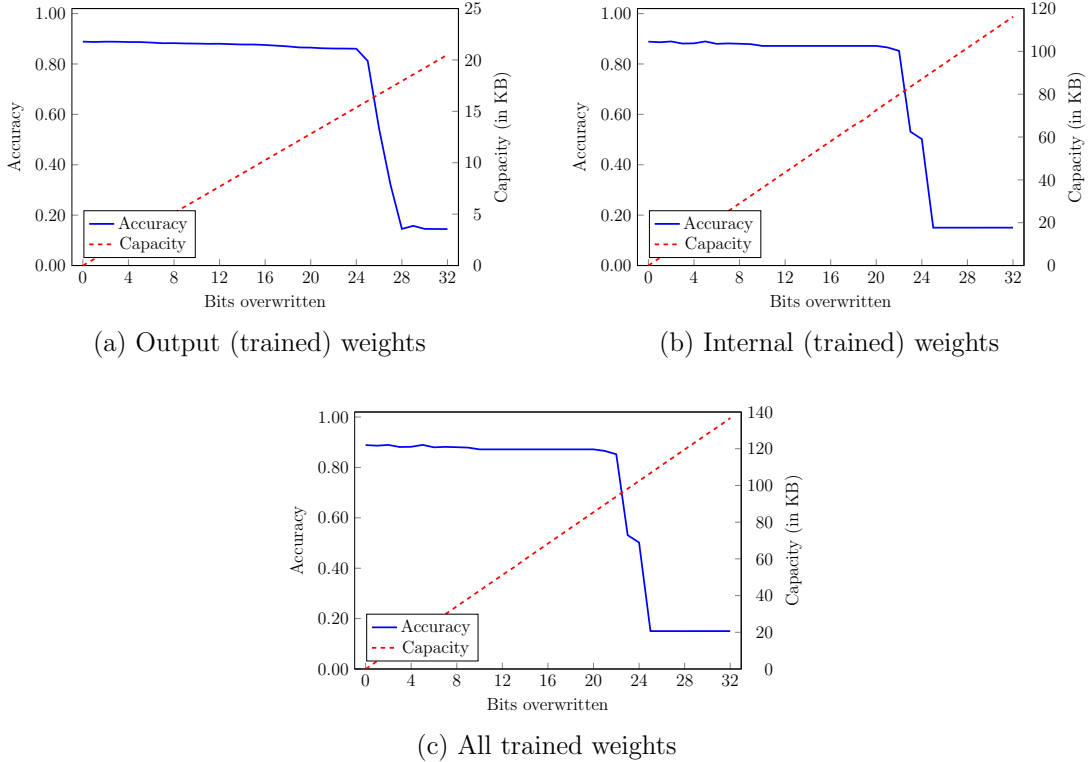

\centering
\begin{tabular}{cc}
\adjustbox{scale=0.925}{
\input figures/xception_output.tex
}
&
\adjustbox{scale=0.925}{
\input figures/xception_internal.tex
}
\\
\adjustbox{scale=0.85}{(a) Output (trained) weights}
&
\adjustbox{scale=0.85}{(b) Internal (trained) weights}
\\ \\
\multicolumn{2}{c}{\adjustbox{scale=0.925}{
\input figures/xception_all.tex
}}
\\
\multicolumn{2}{c}{\adjustbox{scale=0.85}{(c) All trained weights}}
\end{tabular}
\caption{Xception steganographic capacity graphs}\label{fig:xception_Plot}
\end{figure}

Figure~\ref{fig:xception_Plot}(c) provides a summary of our experimental results when the 
low-order bits of all weights are overwritten.
The initial accuracy is about~0.88, and there is a marginal---but inconsistent---decline at small
values of~$n$, with the consistent decline beginning when~21 bits of the trained weights
are overwritten. Thus, we take~$n=20$ as the per-weight steganographic capacity of our 
Xception model, with respect to trained weights.

In this particular Xception model, the hidden layers have~29,046 weights, 
and the output layer contains~5130 weights, for a total of~34,176 trained weights.
Based on a per-weight capacity of~20 bits, we find the steganographic capacity is~85.44~KB
in the trained weights.

\subsection{ACGAN}

For our ACGAN, we use the
\texttt{Sequential} model from \texttt{keras.models} to train the discriminator and generator. 
The discriminator of the trained ACGAN, is then used as the classifier in our steganographic 
capacity experiments. The hyperparameters tested for the model are listed in Table~\ref{tab:acgan}, 
with the selected values in boldface. 
The confusion matrix for our best ACGAN discriminator model appears in
Figure~\ref{fig:confACGAN} in the appendix.
Note that the ACGAN generator plays no role in our capacity calculations, below.

\begin{table}[!htb]
\caption{ACGAN discriminator hyperparameters tested}\label{tab:acgan}
\centering
\adjustbox{scale=0.85}{
\begin{tabular}{c|c}\midrule\midrule
Hyperparameter & Values tested\\ \midrule
\texttt{pad-size} & \textbf{same} \\
\texttt{batch size} &  \textbf{32}, 128, 256 \\
\texttt{max-epochs} & 10000, \textbf{14000} \\
\texttt{random\_state} & 10, 50, \textbf{100} \\ 
\texttt{momentum} & 0.5, \texttt{\textbf{0.8}} \\
\texttt{learning\_rate\_init} & 0.0001 \textbf{0.0002} \\
\texttt{solver} & \texttt{\textbf{adam}} \\ \midrule\midrule
\end{tabular}
}
\end{table}

\begin{figure}[!htb]
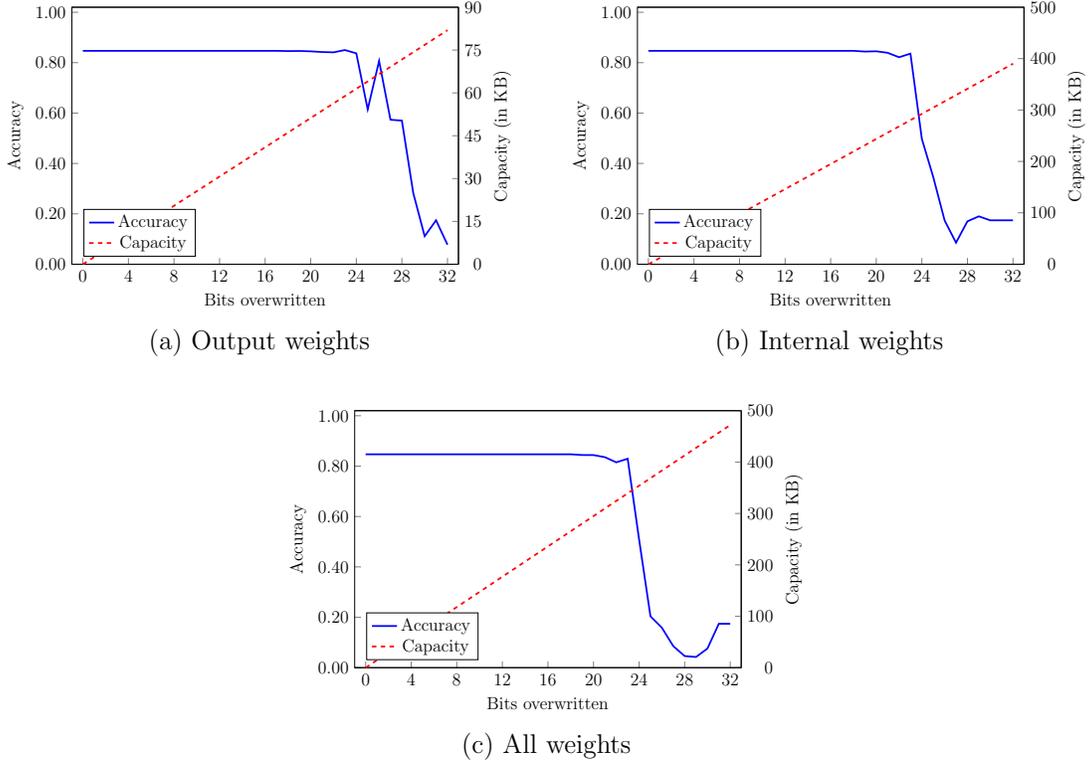

\centering
\begin{tabular}{cc}
\adjustbox{scale=0.925}{
\input figures/gan_out.tex
}
&
\adjustbox{scale=0.925}{
\input figures/gan_in.tex
}
\\
\adjustbox{scale=0.85}{(a) Output weights}
&
\adjustbox{scale=0.85}{(b) Internal weights}
\\ \\
\multicolumn{2}{c}{\adjustbox{scale=0.925}{
\input figures/gan_all.tex
}}
\\
\multicolumn{2}{c}{\adjustbox{scale=0.85}{(c) All weights}}
\end{tabular}
\caption{ACGAN steganographic capacity graphs}\label{fig:ACGAN_plot}
\end{figure}

The results obtained when hiding information in the low-order bits of the weights of our trained discriminator 
model are summarized in Figure~\ref{fig:ACGAN_plot}. 
We observe that the original accuracy for the model is approximately~0.8469, 
and the performance of the model declines by slightly more than~1\%\ when~21 bits
are overwritten, and hence we consider~$n=20$ to be the per-bit capacity.

Figure~\ref{fig:ACGAN_layers} in the appendix gives results the steganographic
capacity results for each of the~4 individual layers in ACGAN. These
graphs follow a similar pattern as the graphs in Figure~\ref{fig:ACGAN_plot}, 
and hence we observe no significant differences between the individual layers.

The ACGAN discriminator has~20,490 weights in the output layer and~97,536 weights in the hidden layers, 
for a total of~118,026 weights. Based on a per-weight capacity of~20 bits,
the total steganographic capacity is~295.065~KB.


\subsection{Discussion}

\begin{figure}[!htb]
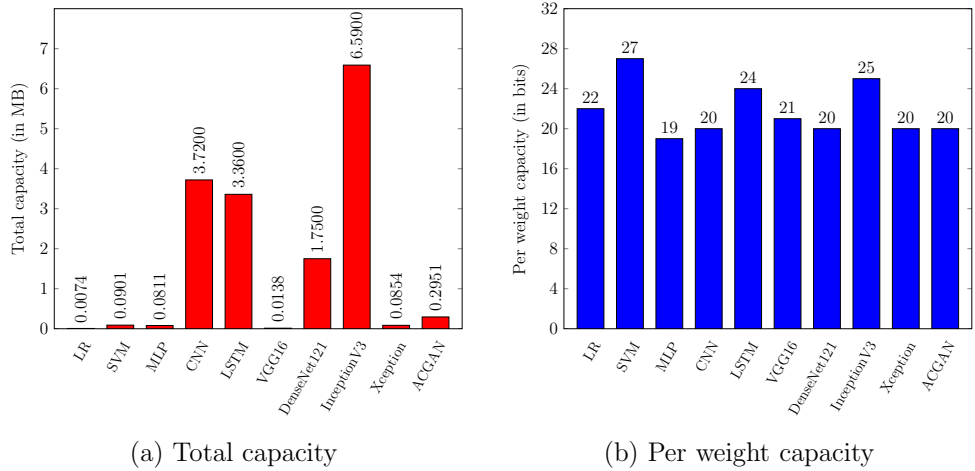

\centering
\begin{tabular}{cc}
\adjustbox{scale=0.6}{
\input figures/barTotal.tex
}
& 
\adjustbox{scale=0.6}{
\input figures/barWeight.tex
}
\\
\adjustbox{scale=0.85}{(a) Total capacity}
&
\adjustbox{scale=0.85}{(b) Per weight capacity}
\end{tabular}
\caption{Bar graphs of capacity results}\label{fig:bar}
\end{figure}

We summarize our steganographic capacity findings in Table~\ref{tab:summary}
and, in bar graph form, in Figure~\ref{fig:bar}.
Of the models tested, SVM has the highest capacity per weight, which implies
that this particular model requires the least precision in its weights.
This is not surprising, given that the SMO algorithm that is used to train SVMs
relies on the fact that low precision suffices. 
Of the pre-trained transfer learning models, InceptionV3 has the highest
capacity per-weight, with respect to trained weights.

\begin{table}[!htb]
\caption{Summary of results}\label{tab:summary}
\centering
\adjustbox{scale=0.85}{
\begin{tabular}{c|c|r|cr}\midrule\midrule
\raisebox{-3.5pt}{\multirow{2}{*}{Model}} & \raisebox{-3.5pt}{\multirow{2}{*}{Layers}} 
	& \raisebox{-3.5pt}{\multirow{2}{*}{Weights\ \,}} 
	& \multicolumn{2}{c}{Steganographic capacity} \\[0.65ex] \cline{4-5}
  & & & Bits per weight & Total\ \ \vphantom{$N^{N^{N^N}}$} \\ \midrule
LR & All & 2560 & 22 & 7.04~KB \\  %
SVM & All & 26,703 & 27 & 90.12~KB \\  %
MLP & All & 34,148 & 19 & 81.10~KB \\  %
CNN & All & 1,489,674 & 20 & 3.72~MB \\  %
LSTM & All & 1,119,626 & 24 & 3.36~MB \\  %
VGG16 & Trained & 5130 & 21 & 13.847~KB \\ 
DenseNet121 & Trained & 700,106 & 20 & 1.75~MB \\ 
InceptionV3 & Trained & 2,107,392 & 25 & 6.59~MB \\ 
Xception & Trained & 34,176 &  20 &  85.44~KB \\ 
ACGAN & Discriminator & 118,026 & 20 & 295.07~KB \\ %
\midrule\midrule
\end{tabular}
}
\end{table}

Note that the numbers in Table~\ref{tab:summary} and Figure~\ref{fig:bar}
for the pre-trained models (VGG16, DenseNet121, InceptionV3, Xception)
only include the trained weights, that is, the weights that were retrained for
the specific malware classification problem under consideration. 
In Section~\ref{sect:V3}, we found
that the per-weight capacity 
for the pre-trained layers of the InceptionV3
model was just~14 bits, as compared to~25 bits for its trained weights.
In spite of this low per-weight capacity, the number of 
pre-trained weights in InceptionV3 is large, and hence the 
steganographic capacity is large---if we consider all weights,
the capacity is~44.74~MB. Similarly, in Section~\ref{sect:VGG} we showed
that if we consider all weights of the VGG16 model, it has also
has an extremely high steganographic capacity at~36.80~MB.

\section{Conclusion}\label{chap:5}

The primary goal of this research was to determine reasonable
lower bounds for the stenographic capacity of a representative sample of 
learning models. Each model was trained on a dataset of more 
than~15,000 malware executables from~10 families, 
with more than~1000 samples per family. 

All of the trained learning models underwent a similar testing procedure:
We first determined the accuracy of a model on the test set, then we embedded information
in the~$n$ low-order bits of the weights, for~$n=1,2,\ldots,32$, and we
recomputed the classification accuracy for each~$n$. 
For generic deep learning models, we experimented with the output layer weights, 
the hidden layer weights, and all of the weights, while for pre-trained
models, we considered the trained weights.
The results were fairly consistent across all models, in that a substantial number of 
bits per weight can be used to hide information, with minimal effect on the accuracy. In addition, 
at some point, the accuracy of all models dropped precipitously, indicating a minimum
level of required precision. These results were also reasonably consistent
across the various layers of the models, with the only notable exception being
the pre-trained weights of the InceptionV3 model, which had a lower
per-weight steganographic capacity.

Our experimental results show that the steganographic capacity of the models 
we tested is surprisingly high. This is potentially a significant security issue, 
since such models are ubiquitous, and hence it
is to be expected that attackers will try to take advantage of them. Embedding, say, malware
in a learning model offers an attack vector that is practical, and could be highly effective in practice.

It would be wise to reduce the steganographic capacity of learning models. Our results indicate
that standard~32-bit weights do not yield a significant improvement in accuracy over what could 
be achieved with, say, 16-bit weights, and for some models, 8-bit weights would be more than sufficient. 

Further research into other popular deep learning models would be worthwhile. 
Also, training models on different types of problems---including classification problems of varying
levels of difficulty---would tell us whether the capacity of a specific model varies
with the difficulty of the problem.
Additional analysis of the pre-trained weights of transfer learning models would be interesting.
Research on compressed models that use smaller numbers of bits to store
each weight would be of practical significance. 

Dropout regularization in, say, MLPs (equivalently, cutouts in CNNs) is used to
force more neurons to be active in training, which can be very effective in
reducing overfitting. It would be interesting to determine whether such regularization 
techniques also affect the precision of trained weights, which can be measured
via the steganographic capacity experiments presented in this paper.

Another area for further investigation
would be to combine some aspects of the steganographic capacity
approach considered in this paper with the work in~\cite{EvilModel}, where
information is hidden in weights that are (essentially) unused by the model.
By combining both of these techniques, we could obtain even larger 
steganographic capacities for learning models.
Finally, it would be interesting---although challenging---to obtain tight
upper bounds on the minimum size of various models,
with the goal of eliminating any usable steganographic capacity.

\bibliographystyle{plain}
\bibliography{references.bib}

\section*{Appendix}

In this appendix, we provide confusion matrices for each of the models analyzed in this paper.
We observe that, in general, \texttt{VB} and \texttt{VBInject} are consistently the most difficult families
to distinguish. We also provide additional steganographic capacity graphs for selected models.

\begin{figure}[!htb]
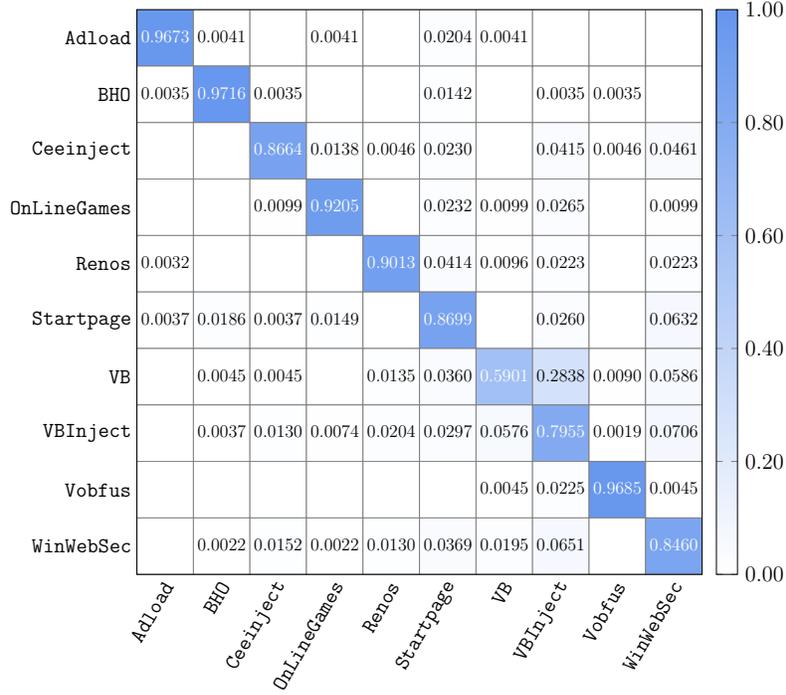

\centering
\adjustbox{scale=0.8}{
\input figures/lr_conf.tex
}
\caption{LR confusion matrix}\label{fig:confLR}
\end{figure}

\begin{figure}[!htb]
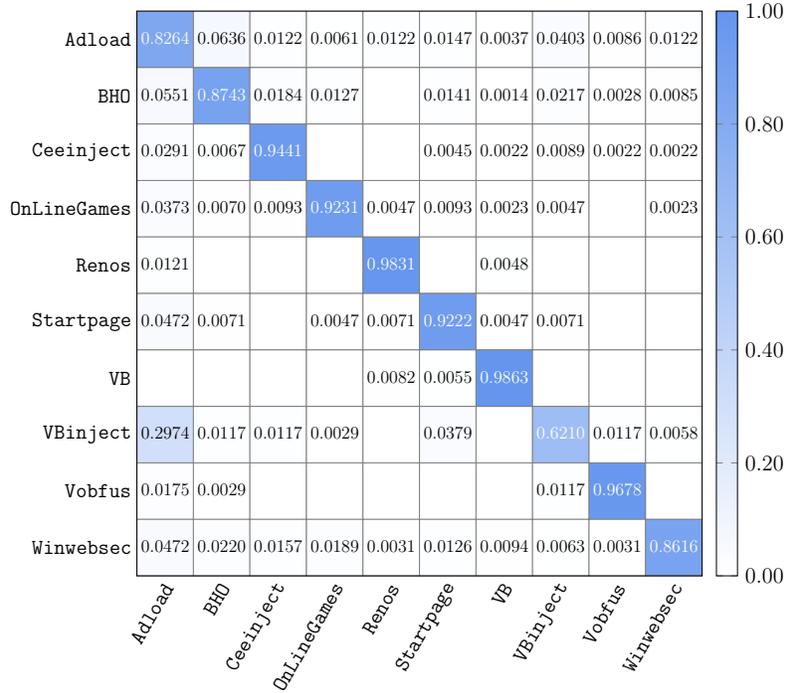

\centering
\adjustbox{scale=0.8}{
\input figures/confSVM.tex
}
\caption{SVM confusion matrix}\label{fig:confSVM}
\end{figure}


\begin{figure}[!htb]
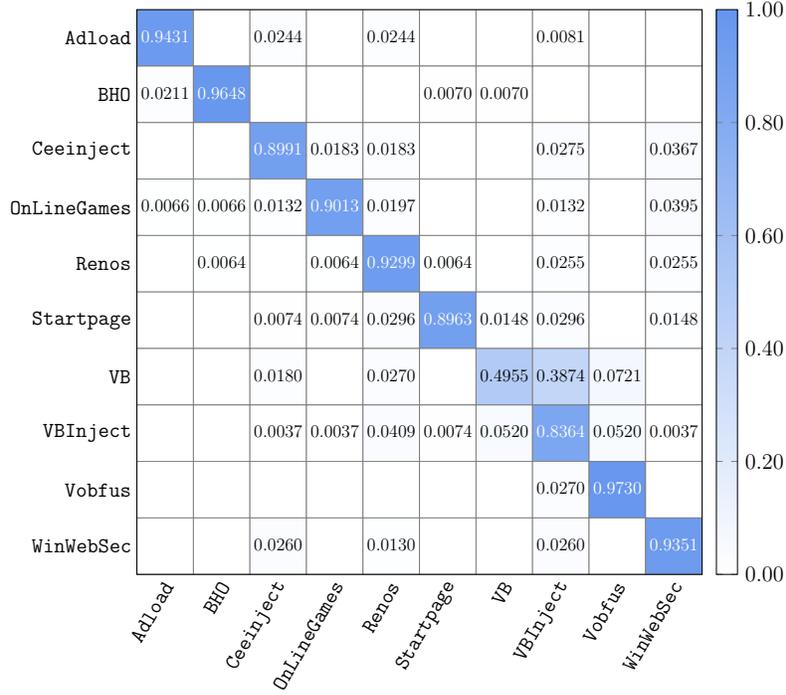

\centering
\adjustbox{scale=0.8}{
    \input figures/cnn_conf.tex
}
\caption{CNN confusion matrix}\label{fig:confCNN}
\end{figure}

\begin{figure}[!htb]
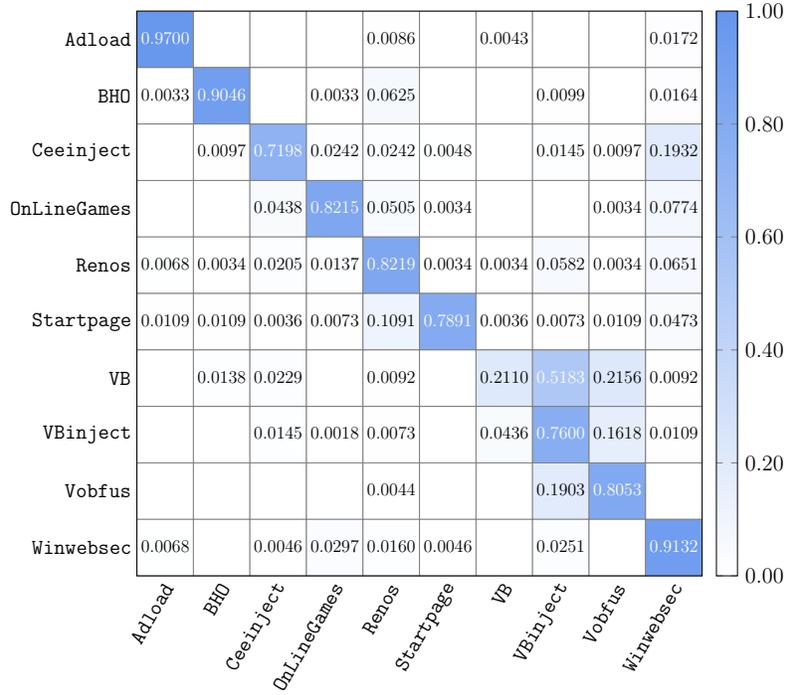

\centering
\adjustbox{scale=0.8}{
\input figures/confLSTM.tex
}
\caption{LSTM confusion matrix}\label{fig:confLSTM}
\end{figure}

\begin{figure}[!htb]
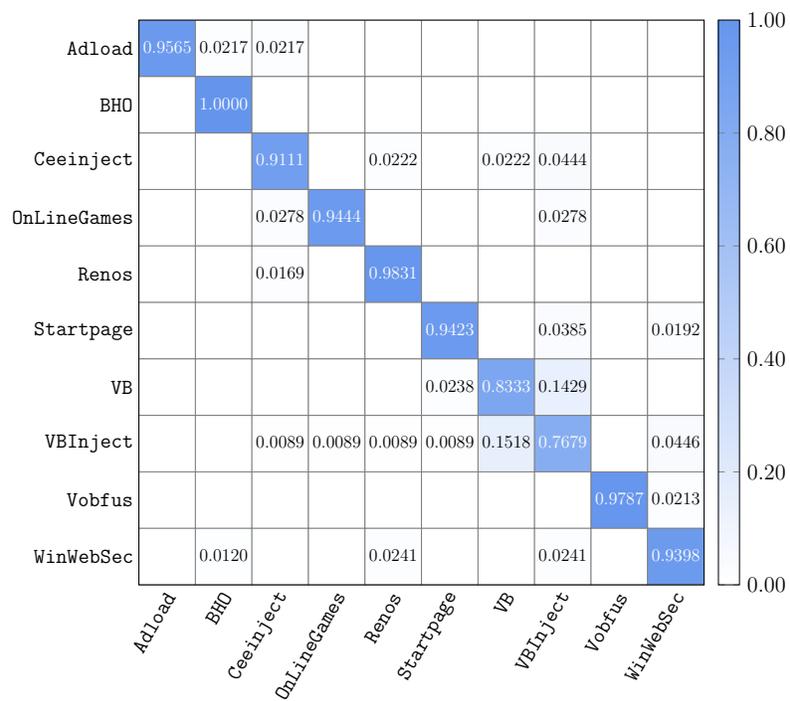

\centering
\adjustbox{scale=0.8}{
\input figures/confVGG.tex
}
\caption{VGG16 confusion matrix}\label{fig:confVGG}
\end{figure}

\begin{figure}[!htbp]
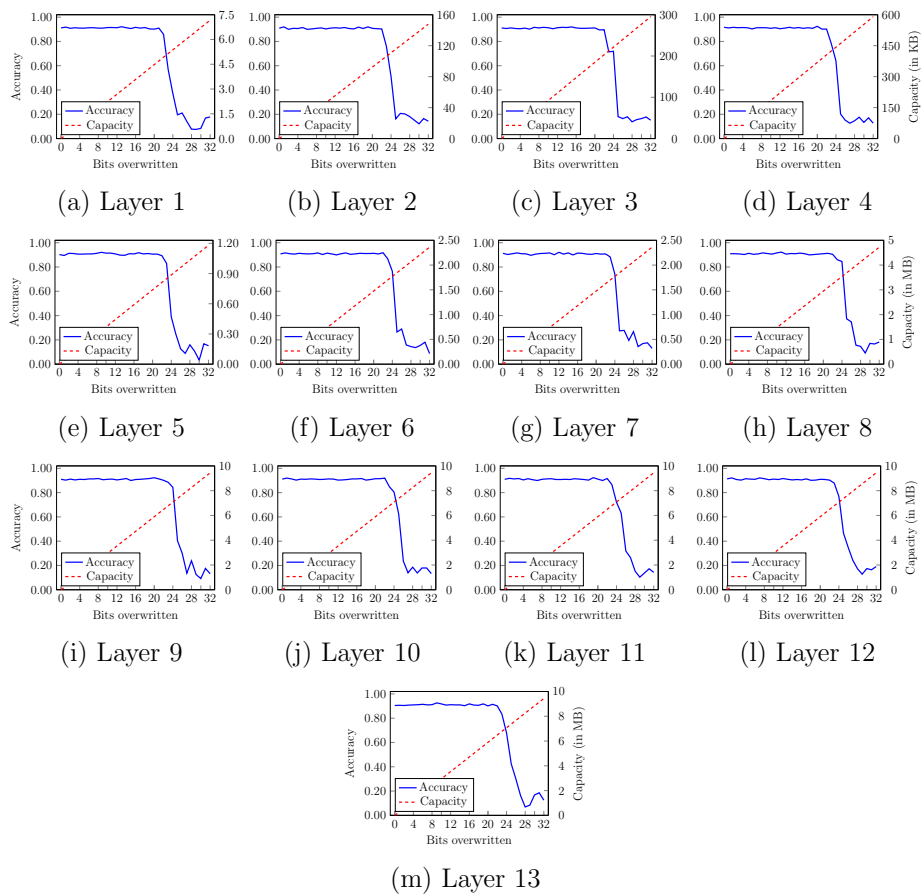

\centering\advance\tabcolsep by -7.0pt
\begin{minipage}{0.75\textwidth}
\begin{tabular}{cccc}
    \adjustbox{scale=0.65}{
    \input figures/layer_1_graph.tex
    }
    &
    \adjustbox{scale=0.65}{
    \input figures/layer_2_graph.tex
    }
    &
    \adjustbox{scale=0.65}{
    \input figures/layer_3_graph.tex
    }
    &
    \adjustbox{scale=0.65}{
    \input figures/layer_4_graph.tex
    }
    \\
    \adjustbox{scale=0.85}{(a) Layer~1}
    &
    \adjustbox{scale=0.85}{(b) Layer~2}
    &
    \adjustbox{scale=0.85}{(c) Layer~3}
    &
    \adjustbox{scale=0.85}{(d) Layer~4}
    \\ \\[-2ex]
    \adjustbox{scale=0.65}{
    \input figures/layer_5_graph.tex
    }
    &
    \adjustbox{scale=0.65}{
    \input figures/layer_6_graph.tex
    }
    &
    \adjustbox{scale=0.65}{
    \input figures/layer_7_graph.tex
    }
    &
    \adjustbox{scale=0.65}{
    \input figures/layer_8_graph.tex
    }
    \\
    \adjustbox{scale=0.85}{(e) Layer~5}
    &
    \adjustbox{scale=0.85}{(f) Layer~6}
    &
    \adjustbox{scale=0.85}{(g) Layer~7}
    &
    \adjustbox{scale=0.85}{(h) Layer~8}
    \\ \\[-2ex]
    \adjustbox{scale=0.65}{
    \input figures/layer_9_graph.tex
    }
    &
    \adjustbox{scale=0.65}{
    \input figures/layer_10_graph.tex
    }
    &
    \adjustbox{scale=0.65}{
    \input figures/layer_11_graph.tex
    }
    &
    \adjustbox{scale=0.65}{
    \input figures/layer_12_graph.tex
    }
    \\
    \adjustbox{scale=0.85}{(i) Layer~9}
    &
    \adjustbox{scale=0.85}{(j) Layer~10}
    &
    \adjustbox{scale=0.85}{(k) Layer~11}
    &
    \adjustbox{scale=0.85}{(l) Layer~12}
    \\ \\[-2ex]
    \multicolumn{4}{c}{\adjustbox{scale=0.65}{
    \input figures/layer_13_graph.tex
    }}
    \\
    \multicolumn{4}{c}{\adjustbox{scale=0.85}{(m) Layer~13}}
\end{tabular}
\end{minipage}
\caption{VGG16 capacity graphs for individual layers}\label{fig:VGG_layers}
\end{figure}

\begin{figure}[!htb]
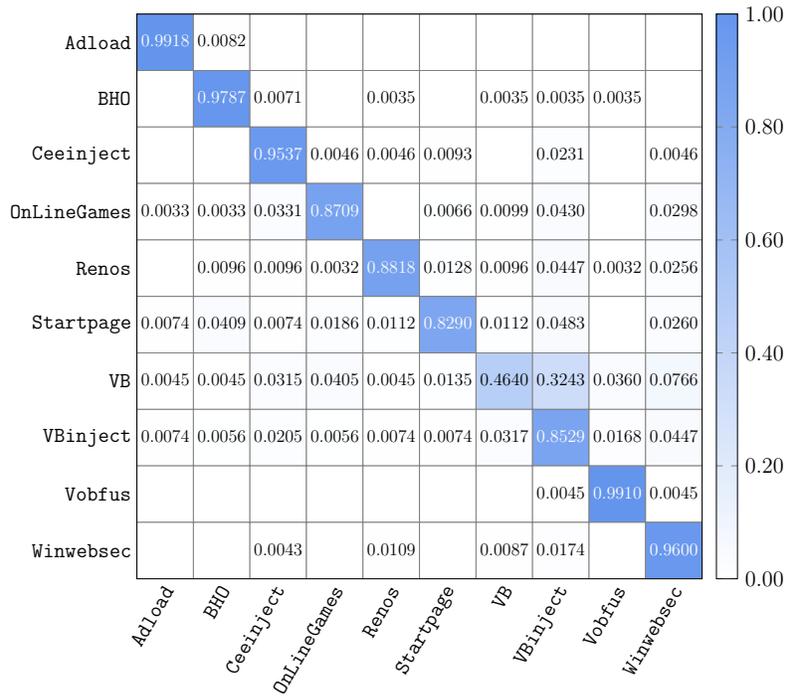

\centering
\adjustbox{scale=0.8}{
\input figures/confDense.tex
}
\caption{DenseNet121 confusion matrix}\label{fig:confDense}
\end{figure}

\begin{figure}[!htb]
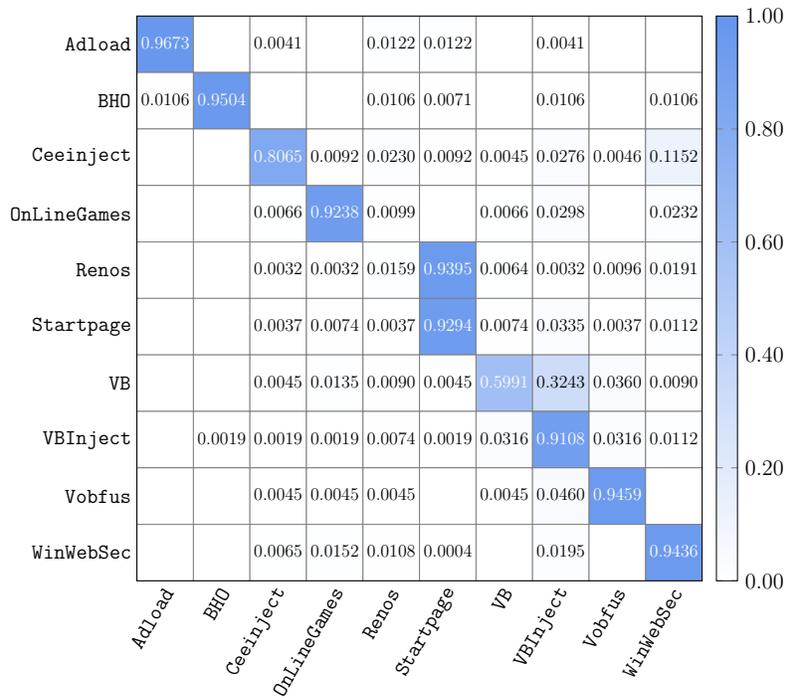

\centering
\adjustbox{scale=0.8}{
\input figures/confInceptionv3.tex
}
\caption{InceptionV3 confusion matrix}\label{fig:confIncept}
\end{figure}


\begin{figure}[!htb]
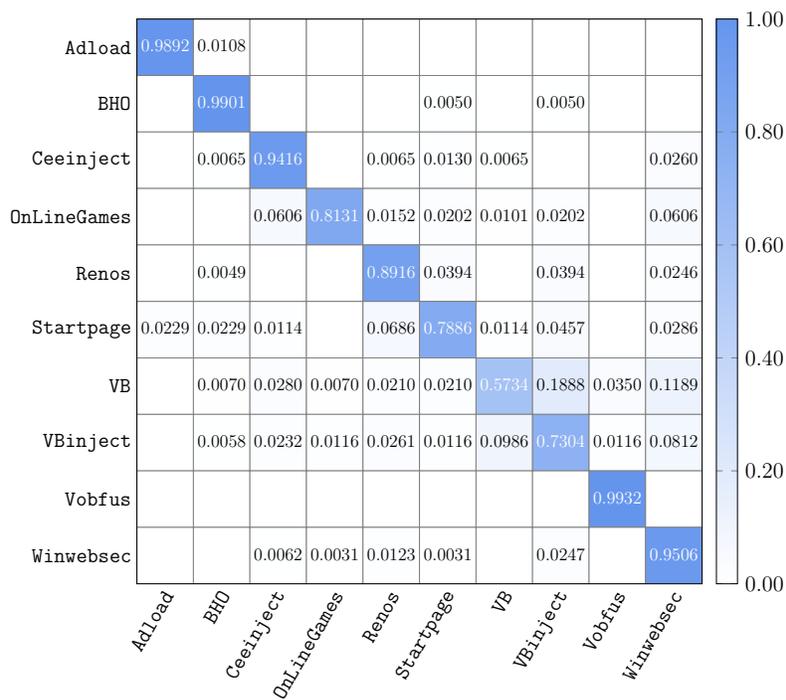

\centering
\adjustbox{scale=0.8}{
\input figures/confXcept.tex
}
\caption{Xception confusion matrix}\label{fig:confXcept}
\end{figure}

\begin{figure}[!htb]
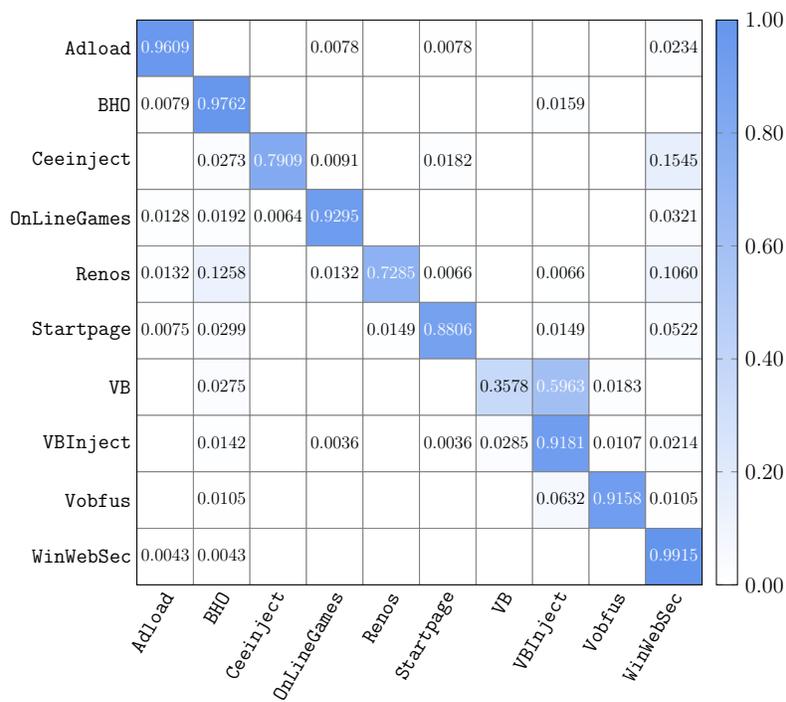

\centering
\adjustbox{scale=0.8}{
\input figures/confACGAN.tex
}
\caption{ACGAN confusion matrix}\label{fig:confACGAN}
\end{figure}

\begin{figure}[!htbp]
\centering
\begin{tabular}{cc}
    \adjustbox{scale=0.85}{
    \input figures/layer_1_ACGAN.tex
    }
    &
    \adjustbox{scale=0.85}{
    \input figures/layer_2_ACGAN.tex
    }
    \\
    \adjustbox{scale=0.85}{(a) Layer~1}
    &
    \adjustbox{scale=0.85}{(b) Layer~2}
    \\ \\[-2ex]
    \adjustbox{scale=0.85}{
    \input figures/layer_3_ACGAN.tex
    }
    &
    \adjustbox{scale=0.85}{
    \input figures/layer_4_ACGAN.tex
    }
    \\
    \adjustbox{scale=0.85}{(c) Layer~3}
    &
    \adjustbox{scale=0.85}{(d) Layer~4}
  \end{tabular}
\caption{ACGAN capacity graphs for individual layers}\label{fig:ACGAN_layers}
\end{figure}

\end{document}